\documentclass{article}

\usepackage[preprint, nonatbib]{neurips_data_2024}

\usepackage[utf8]{inputenc} 
\usepackage[T1]{fontenc}    
\usepackage{hyperref}       
\usepackage{url}            
\usepackage{booktabs}       
\usepackage{amsfonts}       
\usepackage{nicefrac}       
\usepackage{microtype}      
\usepackage[table]{xcolor}       
\usepackage[numbers]{natbib}

\usepackage{soul}
\usepackage{xspace}
\usepackage{graphicx}
\usepackage{listings,multicol}
\usepackage[caption=false]{subfig}  
\usepackage[export]{adjustbox}
\usepackage{amsmath}
\usepackage{amssymb}
\usepackage{textcomp}
\usepackage{listings}
\usepackage{breakurl}
\usepackage{algorithm}
\usepackage[noend]{algpseudocode}
\usepackage{hyperref}
\usepackage{cleveref}
\usepackage{booktabs}
\usepackage{multirow}
\usepackage{enumerate}
\usepackage[shortlabels]{enumitem}
\usepackage{makecell}
\usepackage{spverbatim}
\usepackage{fancyvrb}
\usepackage{tabularx}
\usepackage{tablefootnote}

\newcommand{\ie}{\emph{i.e.,}\xspace}
\newcommand{\eg}{\emph{e.g.,}\xspace}
\newcommand{\secpart}[1]{\smallskip\textbf{#1.}\xspace}

\title{Long Code Arena: a Set of Benchmarks \\ for Long-Context Code Models}

\author{%
  \textbf{Egor Bogomolov$^{1,2}$, Aleksandra Eliseeva$^1$, Timur Galimzyanov$^1$, Evgeniy Glukhov$^1$,}\\
  \textbf{Anton Shapkin$^1$, Maria Tigina$^1$, Yaroslav Golubev$^1$, Alexander Kovrigin$^1$,} \\
  \textbf{Arie van Deursen$^2$, Maliheh Izadi$^2$, Timofey Bryksin$^1$} \\
  $^1$JetBrains Research, $^2$Delft University of Technology\\
  \texttt{lca@jetbrains.com} \\
}

\begin{document}

\maketitle

\begin{abstract}
  Nowadays, the fields of code and natural language processing are evolving rapidly. In particular, models become better at processing long context windows --- supported context sizes have increased by orders of magnitude over the last few years. However, there is a shortage of benchmarks for code processing that go beyond a single file of context, while the most popular ones are limited to a single method. With this work, we aim to close this gap by introducing Long Code Arena, a suite of six benchmarks for code processing tasks that require project-wide context. These tasks cover different aspects of code processing: library-based code generation, CI builds repair, project-level code completion, commit message generation, bug localization, and module summarization. For each task, we provide a manually verified dataset for testing, an evaluation suite, and open-source baseline solutions based on popular LLMs to showcase the usage of the dataset and to simplify adoption by other researchers. We publish the benchmark page on HuggingFace Spaces with the leaderboard, links to HuggingFace Hub for all the datasets, and link to the GitHub repository with baselines: \url{https://huggingface.co/spaces/JetBrains-Research/long-code-arena}.
\end{abstract}

\section{Introduction}\label{sec:introduction}

The Machine Learning for Software Engineering (ML4SE) domain has gained popularity over the recent years, with increasingly more powerful models for text and code processing becoming available. According to a recent survey~\cite{hou2023large}, the most common ML4SE tasks studied in the literature are code generation, code completion, code summarization, and program repair. Unfortunately, the majority of existing benchmarks for assessing ML4SE models have two major limitations: a short length of the available context and a limited resemblance of the practical use cases~\cite{hellendoorn2019code,kovalenko2018does}. 

Two common trends in modern natural language processing (NLP) are retrieval-augmented generation~\cite{gao2023retrieval} and utilization of long contexts~\cite{tay2022efficient}. 
Retrieval-augmented approaches~\cite{borgeaud2022improving,jiang2023active} can base their predictions on information from large corpora of data using various search techniques, while the development of new architectures~\cite{poli2023hyena,fu2024monarch,gu2023mamba} and techniques~\cite{flash-attention-2,bertsch2024unlimiformer} allows models to process tens of thousands or even millions of tokens. 
Both long-context and retrieval-augmented models can in theory utilize information from an entire software project. 
However, most existing ML4SE benchmarks operate with short code snippets --- methods or at most files. 
For example, two most popular code generation datasets---HumanEval~\cite{chen2021evaluating} and MBPP~\cite{mbpp}---require models to comprehend fewer than 1,000 tokens and generate a short function, usually no more than 100 tokens long.

Researchers already do work on extending available context for ML4SE benchmarks. As an example, several recent works investigate code completion at the repository level~\cite{liu2023repobench,repocoder}. However, their usage of software data does not account for the iterative nature of software development: while solving the code completion task in a single file, the benchmarks allow models to use the rest of the project without restrictions. At the same time, other parts of the project can be written after the studied file and utilize its contents, giving the model hints that will not be present in the practical use-case.

In this work, we present \emph{Long Code Arena}, a suite of novel benchmarks for ML4SE models that cover six tasks: library-based code generation, CI builds repair, project-level code completion, commit message generation, bug localization, and module summarization. We design all the tasks and datasets in such a way that they require models to use information from a project module or the entire project to successfully complete the task. For all the tasks, samples used for evaluation are rigorously filtered and then manually verified to ensure the best possible data quality. The data for all the tasks comes from open-source repositories with permissive licenses. We also provide baseline solutions for all the tasks based on popular models, although this work does not aim at solving the tasks --- baselines are provided solely to aid future research.

In the paper, we describe the data collection methodology for each task, describe the evaluation setup, and briefly discuss the implemented baselines. 
At the end of the paper, we also discuss related work and the drawbacks of the existing datasets. 
In the Supplementary materials, we thoroughly describe the structure of the datasets and provide a detailed description of the baselines. 
We open-source the implementations of baselines along with code for evaluation, they can be found on GitHub\footnote{Long Code Arena baselines: \url{https://github.com/JetBrains-Research/lca-baselines}} and serve as an example of using the collected datasets. 
You can access the leaderboard and links to all the datasets (published via HuggingFace Hub) in our HuggingFace Space.\footnote{Long Code Arena leaderboard: \url{https://huggingface.co/spaces/JetBrains-Research/long-code-arena}}
\section{Long Code Arena}\label{sec:methodology}

Long Code Arena is a suite of six benchmarks that cover different aspects of code processing: generation, repair, completion, summarization, processing diffs. For each task, we gather an evaluation dataset of around a hundred to a thousand examples that requires models to operate with source code at the scale of a module or an entire repository. For most tasks, we focus on Python code due to its popularity and to manually verify the correctness of the samples. However, the collection methodology for all the tasks allows extending the benchmarks with more languages in the future. 

All the datasets we collect in Long Code Arena are based on data from open-source GitHub repositories --- source code, commit history, issues, as well as build data from GitHub Actions. 
First, we extract a common corpus of repositories for further processing. 
To do so, we get the list of repositories via GitHub Search~\cite{dabic2021data} that pass the following filters used in other works to ensure the quality of the data~\cite{github2014perils}: at least 1,000 commits, at least ten contributors, issues, and stars, at least 10,000 lines of code, not a fork, last commit after 01.06.2023, and a permissive license (we use the most popular permissive licenses~\cite{vendome2017license} --- MIT, Apache-2.0, BSD-3-Clause, and BSD-2-Clause). 
After the filtering, we are left with 4,343 repositories that we then download via GitHub API along with issues and pull requests data. 
For the CI builds repair task, we also retrieve GitHub Actions logs for some repositories, which we describe further. The only task that we base on the existing dataset is commit message generation, for which we find samples with large commits and long commit messages in the recent CommitChronicle dataset~\cite{our-cmg-paper}.

After the initial data collection stage, we prepare evaluation datasets for each of the six tasks separately. For this, we apply further task-specific filters to the collected data, and then manually examine the samples to ensure their correctness. The following subsections contain in-detail descriptions of all the benchmarks.

\subsection{Library-based Code Generation}\label{sec:code-generation}

\secpart{Task description} 
The first task in our work is a novel library-based code generation task. 
Given a task description and access to the contents of a software library, the model should generate a single file that solves the task heavily utilizing methods from the given library. 
The problem is motivated by the need of programmers to write code that utilizes the present dependencies and in-project APIs rather than adding new dependencies and increasing project complexity.

In contrast to library-based code generation, existing code generation benchmarks require models to produce self-sufficient code snippets, such as solutions to algorithmic problems~\cite{chen2021evaluating,mbpp,apps}, domain-specific code~\cite{card2code}, one-liners~\cite{conala}, etc. 
Among the existing works, the setup of the library-based code generation task is similar to repository-level code completion benchmarks that evaluate API completion~\cite{liu2023repobench,repocoder}. Contrary to them, our benchmark requires models to generate an entire program based on an instruction in natural language instead of a single API call or a single line. 

\secpart{Collection methodology} 
To prepare the benchmark, we first extract usage examples from the Python projects that we collected by finding directories in the project roots that contain ``examples'' in their name. Such usage examples are provided by the library authors in order to show the capabilities and use cases of their libraries. 

After collecting the examples, we filter them: (i) remove examples shorter than 100 or longer than 40,000 characters (excluding comments), (ii) remove examples that have fewer than 400 characters of comments in order to then write high-quality instruction for generation, (iii) remove examples that use fewer than ten API calls specific to the given library. To identify library-specific API calls, we extract names of all functions and classes defined in the mined Python projects and count as library-specific only the ones that appear in a single library. These filters result in 150 files (usage examples) from 62 libraries, with each file heavily relying on the APIs of the respective project.

To create instructions, we first run the selected 150 files through GPT-4~\cite{gpt4}, prompting it to generate an instruction for generating the respective file. This leaves us with step-by-step instructions that the LLM should then follow to generate a script that utilizes the library at hand. Then, we manually fix each instruction in order to reduce hinting to specific library methods and ensure its correctness.

To build contexts for generation, benchmark users have access to contents of the libraries that include on average 254 Python files with 2.5M characters and 2,242 unique class and method names. The respective medians are 164 files, 1.4M characters, and 1,412 names.

\secpart{Metrics} Following the previous work on metrics for assessing code generation quality~\cite{evtikhiev2023out}, we employ ChrF~\cite{chrf} to measure how similar the generated code is to the original human-written one. Additionally, to assess usage of the respective library, we measure \emph{API Recall} calculated as the ratio of library-specific API calls (called functions, instantiated classes, used constants) made by the ground truth solution that also appear in the generated program.

\secpart{Baselines} We evaluate six models: proprietary GPT-3.5-turbo and GPT-4~\cite{gpt4}, and instruction-tuned versions of open-source CodeLlama-7B, CodeLlama-70B~\cite{codellama}, Mistral-7B~\cite{mistral}, and Mixtral-8x7B~\cite{mixtral}. In the first setup, we assess the models' ability to generate code based solely on instruction, without access to the library. In the second setup, we accompany the instruction with 20 method and class names most similar to the instruction according to BM-25~\cite{bm25}. In both setups, GPT-4 shows the best quality with the API Recall of 37\%, while open-source models without library context achieve the API Recall of 7--11\%. BM-25 retrieval allows to improve the API Recall for all models except for GPT-4 by 3--6\%, leaving a huge space for further improvement.

\subsection{CI Builds Repair}

\secpart{Task description} 
The second task in our benchmark suite is fixing failing CI builds. 
This task asks models to generate a patch that fixes a real-life issue in a CI setup. 
The minimal set of data for the task consists of a repository snapshot at the commit that caused the failure of the workflow  (\textit{failed commit} hereafter) and the logs of the failed step. 
The task can also be performed in a simplified \textit{oracle} setup by prompting a model with a list of files and code blocks in them to change. In this case, the code blocks come from the ground-truth fixing diff provided in the dataset.
An important feature of this task is run-based evaluation: we utilize GitHub Actions~\cite{actions} to run the generated fixes and assess their correctness.

\secpart{Collection methodology} To collect the data, we iterate over the 100 largest downloaded Python repositories and get a full list of action runs in each repo started in the last 90 days, as older GitHub Actions logs are not available. The downloaded data contains action status (failed or successful) and links to the action runs. Then, we group actions by branch and workflow names, limiting them to up to three branches per repository and three workflows per branch to ensure data diversity. This way, we get the time-ordered list of actions for each branch-workflow combination. From it, we get a list of pairs of consecutive actions (workflow runs) where the first commit caused a failure of the GitHub workflow, and the next one was successful. Thus, we get a set of failed-success pairs of actions for each branch-workflow pair. We trim the set to three pairs per branch-workflow pair for data diversity.

For each extracted pair of actions, we download logs of the specific failed step of the failed workflow run, the diff between the failed and successful commits, and the meta-information of the failed commit.
We filter out runs that take more than ten minutes, workflows that need tokens/secrets to run, and diffs lacking modifications of code files.
Then, we assess the datapoints, verifying that logs contained all the necessary information to fix the issue, and grade the difficulty of solving datapoints on a 1--3 scale, with 1 corresponding to pure formatting problems, 2 --- local (one-line) errors, 3 --- requiring understanding of the complex file- or project-level dependencies to perform changes in multiple files.

To ensure that the benchmark works as intended, we re-run CI with and without the presumably correct fix that we got during the collection stage. We filter out the workflows that no longer constitute a failed-fixed pair.
Finally, to isolate the problem to a single failure reason, we delete all \texttt{.yaml} files in the \texttt{.github/workflows/} directory except for the failing workflow. 

The total size of the final dataset is 77 items: 35 with difficulty 1, 14 --- with difficulty 2, and 28 --- with difficulty 3. The median length of the logs is 6.5K symbols with an average of 145K symbols due to a few extremely long logs. The mean and median for the number of files in the repositories is 610 and 240, for the number of lines --- 170K and 56K, number of symbols --- 7.5M and 2.4M.

\secpart{Evaluation}
We provide the code for evaluation in our repository with baselines.\footnote{Code for running evaluation of CI builds repair: \url{https://github.com/JetBrains-Research/lca-baselines/tree/main/ci-builds-repair/ci-builds-repair-benchmark}} 
After a model generates a patch for fixing the build, the benchmark uploads it to a separate branch in the forked GitHub repository and runs a CI workflow there. 
Then, it collects the results of the CI run, allowing us to compute the number of resolved runs and to check the arising mistakes. 
The target metric for the CI builds repair task is the percentage of successfully fixed builds.

\secpart{Baselines}
We run several LLMs on the CI builds repair benchmark. 
We use an \textit{oracle} setup for the baselines, prompting the models to change the code blocks that were edited in the ground-truth fixing diff.
To pass context from the build logs, we find the first occurrence of the case-insensitive substring ``error'' in the logs and take a seven-line context around this occurrence (three lines before and after). If the substring is not found, we pass seven last lines of the log. The instruction then reads as follows: ``\textit{Fix CI in order for tests to pass. Relevant logs: \{relevant\_logs\}}''.
We prompt the LLM to modify these code sections to align with the given instructions and pass all the sections in a single request. 
The LLM replies with the edited versions of the code sections that are converted into a diff and returned to the benchmark. 
The results for open-source models such as Mistral-7B~\cite{mistral} and various versions of CodeLlama-Instruct~\cite{codellama} range from 4\% to 9\% of successful fixes, while GPT-3.5 is able to resolve 17\% of samples.

\subsection{Project-Level Code Completion}\label{sec:code-completion}

\secpart{Task description} 
The next task in the suite is project-level code completion, for now targeting the completion of single lines. 
We formulate the task as follows: given relevant information from the project, which we call \emph{context}, and a prefix of the \emph{completion file}, one needs to generate the next line in this file. 
While there exist other repository-level completion datasets~\cite{repocoder,liu2023repobench}, we use project history from Git to mimic the real-world use case and avoid possible data leakages between files that arise when files in the context are written after the completed file and rely on the completed code.
On top of that, we introduce a fine-grained classification of the completed lines by the used APIs.

\secpart{Collection methodology} To create the dataset, we process the collected Python projects, traversing their Git histories to collect commits that were done after 01.01.2022. We extract newly added files from them, filtering out files with fewer than 200 lines or more than 2,000 lines. To collect the context for each file, we checkout the respective parent commit and save the contents of all the code and text files (\eg build files, documentation), constituting the repository as it was when the commit was made. Each datapoint contains the file for completion, a list of lines to complete with their categories (see the categorization below), and a repository snapshot that can be used to build the context.

We split our dataset into four parts based on the total size of \texttt{.py} files in the repository snapshot. As the reference for such a division, we chose the CodeLlama model~\cite{codellama}, which has a context window of size 16K and about three characters per token. 
Based on this, we have four sets of samples with the following limits on the total number of characters in the context \texttt{.py} files: \emph{small-context set} from 0 to 16K $\times$ 3 = 48K characters; \emph{medium-context set} from 48K to 192K characters; \emph{large-context set} from 192K to 768K characters; \emph{huge-context set} from 768K characters. 
We downsample datapoints to five datapoints per repository, and the repositories to 75 per set to ensure data diversity. 
The sizes of the four sets are 144, 224, 270, and 296 datapoints, respectively.

For each datapoint, we also provide a list of lines for completion---35 lines on average---since evaluating a code model on every line of a file is extremely resource-consuming. 
Moreover, not all lines are equally hard to complete; \eg function declaration lines can be challenging due to uncertainty, whereas loop definition can be straightforward. 
Taking that into account, we introduce a classification of the code lines into six categories depending on the used functions and classes. 
Categories \textit{committed}, \textit{inproject}, and \textit{infile} refer to where the used functions/classes are defined: in the same commit as the completed file, in the project snapshot before the commit, or right in the completed file. 
\textit{Common} category is assigned to the lines that contain common functions such as \texttt{main} or \texttt{get}. 
We classify lines as \textit{non-informative} if they are too short, too long, contain prints, etc. (see Appendix~\ref{sec:sheet3} for the full definition), and assign the \textit{random} category to the rest of the lines.

While each line can fall into multiple categories based on the content, we only assign the ``most difficult'' category to each line in the following order (from difficult to easy): \textit{committed}, \textit{inproject}, \textit{infile}, \textit{common}.
We then sample on average ten completion lines per datapoint for informative classes and five lines per datapoint for non-informative and random classes.
Thus, for each file in the dataset, we have multiple lines that the model should complete. Total numbers of completion lines are 4,686, 8,676, 9,631, and 9,810 for each of four sets, respectively.

\secpart{Metrics} The main metric for the project-level code completion task is the exact match of generated lines per category. This is a proportion of correct predictions calculated separately for each of the categories. The prediction is correct if it matches the ground truth after removing leading and trailing whitespaces from both.

\secpart{Baselines} 
We evaluate CodeLlama-7B~\cite{codellama} and DeepSeek Coder of sizes 1.3B and 6.7B~\cite{deepseek-coder}. 
For each model, we evaluate several strategies for composing the input context from the repository files (see Appendix \ref{sec:sheet3} for details). Among them, building the context from files closest in the file tree to the target file works best. 
The boosts for Exact Match for such a context composer for CodeLlama-7B on the medium context set are +16\% for the \textit{infile} lines and +53\% for the \textit{inproject} lines compared to using only the target file as context.
\subsection{Commit Message Generation}

\secpart{Task description}
The fourth benchmark that we present is commit message generation (CMG) for large commits. 
In CMG, a model should generate a natural language description of changes performed in a single commit. 
The changes can be represented in different ways --- in various diff formats, as separate versions of each file before and after the changes took place, and others. 
Moreover, models can utilize information from unchanged project files to better understand how changes impacted the project.
CMG is a well-established task in academic research~\cite{cmg-survey} and a prominent feature in developer tools~\cite{jetbrainsCMG,githubCMG}, however, researchers often limit the scope to short diffs~\cite{our-cmg-paper}, leaving the performance on larger commits unexplored. Moreover, the quality of commit messages from open-source repositories---the most common data source---is notoriously mixed~\cite{what_makes_a_good_msg}. 
We bridge these two gaps with our novel CMG benchmark, manually curated and tailored for larger commits.

\secpart{Collection methodology}
We use the CommitChronicle~\cite{our-cmg-paper} dataset as the main data source. 
As the dataset aligns with our needs, we chose to use it rather than rebuild a dataset of commits from scratch from the repositories of Long Code Arena. 
CommitChronicle is a large-scale dataset with 10.7M commits from permissively licensed GitHub repositories in 20 programming languages. 
Notably, CommitChronicle omits restrictive data filtering steps, such as strict limits on the maximum length of code changes, thus fitting perfectly for our use case that targets larger diffs. 
As we are building a benchmark, we use only the \textit{test} subset of CommitChronicle. 
To make manual filtering possible, for now, we limit the work to the Python language and thus consider only the subset of the test set that includes changes to at least one \texttt{.py} file. This results in 172K commits from 455 repositories. 

With CommitChronicle encompassing a wide array of commits, we follow the best practices from previous works~\cite{our-cmg-paper, octopack, starcoder} to filter data and reduce the number of low-quality samples. 
Filtering criteria include minimum length in words and lines, message format, presence of hashes (for the exact criteria, see Appendix~\ref{sec:sheet4}). 
After the filtering, we retain 3,260 commits. 
Since we aim to target commits with larger changes, after the initial filtering, we only keep samples where the number of characters in diffs related to \texttt{.py} files is $\geq$ 3,000 characters. 
This leaves us with 858 commits that we further review manually to keep only those where the commit message provides a comprehensive description of all changes without introducing any external information and the changes are meaningful and non-trivial.

After manual filtering, our resulting dataset comprises 163 commits from 34 repositories. 
We follow the same format as the original CommitChronicle~\cite{our-cmg-paper} dataset and include commit diffs, commit messages, and relevant metadata that allows tracing each commit back to GitHub. 
To facilitate further experiments with constructing context for the CMG task, we provide the sources for all the repositories. 
Diffs in the dataset comprise from 67 to 800 lines, or 3.3K to 41K characters. When taking the full content of the changed files and other project files into account, the context spans from 4K to 5M lines, or 144K to 156M characters.

\secpart{Metrics}
We employ metrics used in previous works, including BLEU~\cite{bleu}, ROUGE~\cite{rouge}, ChrF~\cite{chrf}, and BERTScore~\cite{bertscore}. 
For BERTScore, we additionally include the normalized scores as proposed by the authors of the original metric~\cite{bertscore_normalization} to allow for easier interpretation.

\secpart{Baselines}
We evaluate a range of proprietary and open-source models, including multiple OpenAI models, Mixtral-8x7B~\cite{mixtral}, Mistral-7B~\cite{mistral}, variations of DeepSeek Coder~\cite{deepseek-coder}, versions of CodeLLaMA~\cite{codellama}, and fine-tuned CodeT5~\cite{codet5}. GPT-4 Turbo shows the best results with ChrF of 34.4. The best performing open-source model is Mixtral-8x7B with ChrF of 32, followed by Mistral-7B. 
\subsection{Bug Localization}

\secpart{Task description}
The next problem addressed by the proposed benchmark is the bug localization task. 
This problem can be formulated as follows: given an issue with a bug description and a repository snapshot in a state where the bug is reproducible, identify the files within the repository that need to be modified to address the reported bug.
Although this is a subset of the larger bug-fixing problem, partially covered by SWE-Bench~\cite{jimenez2023swebench}, bug localization requires its own separate evaluation. This independent assessment can provide a better understanding of the various approaches and their efficiency in identifying the precise location of bugs within the large code bases.

\secpart{Collection methodology}
To build the dataset for the bug localization task, we process the previously collected 8M issues and 7M PRs from GitHub with more than 34.4M comments.
The provided issue data contains issue descriptions and labels (\textit{e.g.}, ``bug''), by which we can determine the reason behind creating the issue. 
For pull requests, we extract code diffs and link them to issues they resolve.
We use regular expressions to parse links in PRs' titles, description comments, as well as issue comments (\textit{e.g.}, ``fixes \#24'' or ``\#25 resolved'').

We filter the data to ensure data quality and limit the subset to programming languages familiar to the authors for manual labeling (see Appendix~\ref{sec:sheet5} for the exact procedure). 
After this, we are left with 7,479 pairs of bug issues and pull requests linked to them. 
Out of them, 4,339 modify Python files, 2,522 --- Java files, and 618 --- Kotlin files.
For each language, we manually examine a subset of datapoints to see that they meet the following criteria: the issue description is complete and fully describes the introduced changes, while the changes do indeed fix the issue and do not produce code irrelevant to it. 
Since manual labelling of the entire dataset of 7,479 samples is very time-consuming, we carry out the following procedure. 
For each language---Python, Java, and Kotlin---we manually examine samples iterating over the repositories from the most starred to the least starred, and stop after selecting 50 good datapoints per language. 
Importantly, for the initial set of 7,479 PRs, the median number of changed files is one.
Given that, we select half the samples from fixes that only touch a single file, and half the samples from fixes that change from two to ten files.
In terms of the context size, the median number of files in the repository is 331, with an average of 1K files. Each file typically contains 1.5K tokens. Additionally, issues within the repository generally consist of approximately 150 words, equating to around 400 tokens.

\secpart{Metrics}
The task of bug localization is similar to information retrieval, so we use common metrics from this domain: recall at \texttt{k} (\texttt{R@k}), precision at \texttt{k} (\texttt{P@k}), F1 score (\texttt{f1-score}), and mean average precision (\texttt{MAP}). 
We select \texttt{k} to be equal to 1 for changes that require modification of a single file, and 2 for the rest of the changes.
We compute metrics for these two cases separately and report both.

\secpart{Baselines}
First, we evaluate several retrieval-based approaches: TF-IDF, embeddings from CodeT5~\cite{wang2023codet5plus} and CodeBERT~\cite{codebert}, embedding models GTE~\cite{li2023general} and Mistral~\cite{SFRAIResearch2024}.
We use cosine distance between vectors for ranking. 
Furthermore, we evaluate BM25~\cite{bm25} retrieval provided by llama-index~\cite{Liu_LlamaIndex_2022}. 
GTE model demonstrates the best result with 0.33 \texttt{MAP}, followed by Mistral with 0.3, and TF-IDF with the BPE tokenizer~\cite{DBLP:journals/corr/SennrichHB15} with 0.27. The results for the rest of the models are lower than 0.25 \texttt{MAP}.

We also evaluate two chat models --- GPT-3.5 and GPT-4. 
We prompt them to indicate from one to five bugged files using the issue description and the full list of files from the repository.
If the resulting prompt does not fit into the context size, we split the file list into several queries, followed by the final one that combines all outputs and asks to finalize the result. 
These approaches show better scores compared to the retrieval-based approaches, with GPT-4 achieving 0.39 \texttt{MAP}.

\subsection{Module Summarization}

\secpart{Task description}
The last benchmark in the suite is dedicated to the task of summarizing project modules into natural language.
We formulate the module summarization task as follows: based on the module's or project's source code and intent (a one-sentence description of the expected documentation content), the model should write its textual documentation. 
This task greatly increases the context size available to the models compared to the existing benchmarks that cover method- or class-level summarization~\cite{husain2020codesearchnet,lozhkov2024starcoder, luo2024repoagent}.
The source of inspiration for the module summarization task is the fact that large projects often include high-level materials, such as quick start guides, tutorials, module documentation, and usage instructions. The task aims to alleviate the time-consuming and routine process of creating these materials.

\secpart{Collection methodology}
To collect the dataset, we gather documentation files---files with extensions \texttt{.md}, \texttt{.txt}, and \texttt{.rst}---that are located in the \texttt{docs} directory from the collected Python repositories. 
We then identify the associated code for each file by parsing the documentation and extracting links to files and directories with source code. 
Associated code files can encompass the entire project, particularly for quick-start documentation, or specific files for narrower cases. 
Searching for relevant code is essential to prevent the inclusion of text documents not related to specific parts of the source code, such as installation guides. 
If a file does not correspond to any module, we skip it. 
Subsequently, we remove documents that are fewer than ten lines of text without considering markup (\ie in plain text format).
After the filtering steps, we are left with 461 files.

If a file passes all automatic filters, we review it manually before including into the dataset, ensuring that the text summarizes source code, and the other way around --- information from source code is sufficient to write the documentation. This resulted in the final dataset of 216 files from 43 repositories, for each of which we manually specify an intent based on the documentation headers and contents. 

For each datapoint, we attach the relevant context that was automatically extracted, as well as all the code from the repository with documentation files excluded. This enables researchers to experiment with different context collection techniques.
The average length of the target documentation is 2,549 tokens (8,807 characters). The average length of the code context greatly depends on the sample and can be very large, as sometimes the context might include the code of the entire repository. Thus, the datapoint with minimum length of the relevant code context has 327 tokens, while the average and median length of code context are 18,572 and 21,286 tokens, respectively.

\secpart{Metrics} \label{m2t_metric}
Previous work~\cite{roy2021reassessing} shows that out of n-gram-based metrics, the ChrF metric~\cite{chrf} works best for code summarization tasks. However, it was assessed for short texts, and our experience shows low sensitivity when discriminating long files.  To overcome this limitation, we propose using LLMs as scalable proxies for human assessors, similar to the work of Chiang and Lee~\cite{chiang2023large}.

Our proposed metric \emph{CompScore} feeds an LLM the relevant code and two versions of documentation: the gold standard and the model-generated text. The LLM estimates the probability of one documentation better explaining and fitting the code than the other. To mitigate potential ordering effects in model responses, CompScore calculates the probability that the generated documentation is superior by averaging the results of two queries, swapping the order of the generated and reference documentation. The CompScore ranges from 0 to 100, with ground truth documentation receiving 50.

This scoring method not only provides a robust measure of documentation quality but also incorporates the flexibility and semantic evaluation capacity of human judgment. We use a local instance of Mistral-7B~\cite{mistral} with a greedy generation algorithm to make the evaluation both cost-efficient and reproducible across various computational environments.

\secpart{Baselines}
We conduct all our experiments within a zero-shot setting. For every distinct sample, the model uses information about the target file name, intent, and the code we consider relevant (truncated to the supported length in tokens). We then compare the generated documentation with the ground truth provided in the dataset.
We evaluate a range of proprietary and open-source models, including multiple OpenAI models, versions of Mixtral~\cite{mixtral}, Mistral~\cite{mistral}, CodeLLaMA~\cite{codellama}, and LLaMA~\cite{touvron2023llama}. GPT-4 shows the best results with the CompScore of 57.3. The best performing open-source model is Llama2-70B with 48.2, followed by Llama2-13B and Mistral-7B-v0.3.

\section{Related Work}\label{sec:related-work}

While there exist plenty of ML4SE datasets and even benchmark collections~\cite{codexglue}, most of them require models to operate with rather short contexts, around the size of a single method, which hinders the evaluation of novel long context models. 
Code generation datasets~\cite{chen2021evaluating,mbpp,evalplus,apps,cruxeval,conala} require models to process up to several paragraphs of the problem statement and then generate a short program (one line to one file). 
Existing datasets for code summarization~\cite{husain2020codesearchnet,codexglue} target documentation in a single method, meaning that both input and output size are below several hundred tokens.
Previously developed commit message generation benchmarks~\cite{cmg-survey, our-cmg-paper, commitbench} contain significantly shorter messages and diffs compared to Long Code Arena. 

For code completion, recently, researchers introduced two benchmarks that operate at the repository scale: RepoEval~\cite{repocoder} and RepoBench~\cite{liu2023repobench}, also focusing on the completion of a single line. Compared to these benchmarks, we introduce a fine-grained classification of the completed lines and prevent possible data leakages by traversing Git history. 

SWE-bench~\cite{jimenez2023swebench} is a recent benchmark that requires models to fix issues in real-world programming projects. Long Code Arena covers a more diverse set of tasks, the most similar being CI builds repair, which focuses on builds in general rather than tests, and bug localization, which is a sub-task of the SWE-bench objective that we evaluate on a broader set of languages: Python, Java, and Kotlin.

The most notable benchmarks for long context models include Long Range Arena~\cite{tay2020long} and Scrolls~\cite{shaham2022scrolls}. Our work builds the first such benchmark focusing on ML4SE tasks, while Long Range Arena includes synthetic problems and Scrolls focuses on natural language processing.
\section{Limitations and Future Work}

In order to gather benchmarks for Long Code Arena, we had to make several design decisions that can impact the generalizability.
First, we base the benchmarks on open-source data. 
This allows researchers to experiment with various context-collection techniques because they have access to source code data.
On the other hand, modern LLMs use most available open-source data for training, and such reliance can lead to data contamination, which in turn can skew the evaluation results.

We argue that the tasks that we choose are less prone to models memorizing training data: there is no direct link between answers to benchmark tasks and raw repository data that modern models use for training. 
For example, while models could have seen documentation of specific libraries during training, currently it is unlikely that it was present side by side with the source code of the respective modules.
The most memorization-prone task in our suite is code completion, but for it, we use historic data from Git repositories, which may become changed or overridden by the moment LLMs' training data is scraped.

In order to allow for manual examination of the collected data and to keep the benchmarks consistent, for most tasks we focus on datasets of Python code. 
Fortunately, the data preparation pipeline for all the tasks can be reused to produce datasets for other languages. 
The most complex step in this case will be manual verification and filtering of the data to ensure quality and correctness. 
In order to meet the quality requirement, we leave extension of datasets to other languages for future work.

In addition to extending datasets to other programming languages, future work includes collecting data for fine-tuning models for particular tasks and evaluating more models on the benchmarks. 
In order to assist other researchers with the latter, we open-source the code for the baseline solutions.

\section{Conclusion}

In this paper, we present the \emph{Long Code Arena}. The goal of this work is to stimulate research in ML-based solutions for realistic software engineering tasks.
In particular, we design a series of tasks that require taking a complex context into account, such as full projects, libraries and their usage, and coarse-grained components. 
Our work presents six benchmarks related to code generation, repair, completion, and summarization. 
For each task, we carefully design and manually curate evaluation data, metrics for assessing the results, and baseline solutions based on the pre-trained models.
Our experiments show that the tasks are within reach, but far from solved.
We hope and expect that our Long Code Arena will encourage researchers in ML4SE and NLP communities to advance the field of ML-enabled software engineering.

\bibliographystyle{plainnat}
\bibliography{paper}

\newpage

\appendix

\newcommand{\qII}{This dataset is created by the JetBrains Research team, in particular, by the authors of this paper.}

\newcommand{\qIII}{This work was conducted at JetBrains Research and therefore was funded by JetBrains, a vendor of specialized development tools.}

\newcommand{\qVII}{The dataset is a sample.\xspace}

\newcommand{\qXI}{All instances are independent, yet may share properties such as the same contributor or repository, which are represented as fields in the dataset.}

\newcommand{\qXII}{The dataset only contains data for evaluation (\textit{i.e.}, testing split).}

\newcommand{\qXV}{This dataset was collected from openly available GitHub repositories with permissive licenses, with the assumption that any data found was intended to be shared freely. However, it is possible that these repositories contained confidential materials. The data in the dataset was manually evaluated, and we did not see anything that could be considered confidential.}

\newcommand{\qXVI}{The data comes from GitHub, and hence must comply with GitHub's acceptable use policy, in particular concerning user safety. We also manually verified our data and did not find any violation.}

\newcommand{\qXVII}{The dataset consists of code and artifacts collected from GitHub, meaning that they were written by human users. However, these human users themselves, their coding style, authorship information, authorship of source code in any other way, or personal information in any other way are not the focus of the dataset directly.}

\newcommand{\qXVIII}{We do not provide any markers of subpopulations, since people are not the focus of the dataset. However, some indicators might be possible to deduce by following individual datapoints to their source.}

\newcommand{\qXIX}{The data was collected from GitHub and thus might be traced back to GitHub users.}

\newcommand{\qXX}{This dataset was collected from openly available GitHub repositories with permissive licenses, with the assumption that any data found was intended to be shared freely. However, it is possible that these repositories contained sensitive materials. The data in the dataset was manually evaluated, and we did not see anything that could be considered sensitive.}

\newcommand{\qXXV}{The data collection process was conducted by the authors of this paper.}

\newcommand{\qXXVIII}{The dataset consists of code and artifacts collected from GitHub, meaning that they were written by human users. However, these human users themselves, their coding style, authorship information, authorship of source code in any other way, or personal information in any other way are not the focus of the dataset directly.}

\newcommand{\qXXIX}{We collected the data from GitHub, a website hosting code and artifacts written by humans.}

\newcommand{\qXXX}{Individuals were not notified about the data collection, however, we made sure to only collect the data with permissive licenses, ensuring that it can be reused.}

\newcommand{\qXXXI}{We did not ask for consent directly, however, we made sure to only collect the data with permissive licenses, ensuring that it can be reused. We made sure our data collection procedure is in line with GitHub's acceptable use policies.}

\newcommand{\qXXXII}{On our HuggingFace Space, we provide information on how individuals can request removals.}

\newcommand{\qXXXIII}{Since individuals are not the focus of our dataset, we foresee at most limited impact. Users of the dataset might attempt to trace back artifacts to individuals (via GitHub) and try to reach out to them (via contact information on GitHub) with questions about their artifacts.}

\newcommand{\qXXXVII}{The code for preprocessing is available on demand by contacting the authors.}

\newcommand{\qXL}{The dataset is currently used in our repository with baselines available on GitHub.}

\newcommand{\qXLII}{Not in the data itself. As per the GitHub acceptable usage requirements, researchers using this dataset must make any papers resulting from it available as open access.}

\newcommand{\qXLIII}{To the best of our knowledge, no.}

\newcommand{\qXLV}{Yes, the dataset is publicly available on the internet.}

\newcommand{\qXLVII}{The dataset is already publicly available.}

\newcommand{\qXLVIII}{Data coming from GitHub will be re-distributed under the license it was distributed with originally on GitHub (for which we only used permissive licenses). The terms of use require that research conducted with this dataset makes any resulting paper available as open access, in line with GitHub's requirements.}

\newcommand{\qL}{To the best of our knowledge, no.}

\newcommand{\qLII}{The dataset will be maintained by the JetBrains Research team.}

\newcommand{\qLIII}{The dataset curators can be contacted via email at \texttt{lca@jetbrains.com}.}

\newcommand{\qLIV}{There is no erratum as of June 2024.}

\newcommand{\qLV}{The dataset will be extended to more languages and samples over the course of time.}

\newcommand{\qLVI}{On the HuggingFace Space, we provide information on how individuals can request removals.
}

\newcommand{\qLVII}{The older versions will be kept around for consistency.}

\newcommand{\qLVIII}{We welcome all contributions and encourage others to contact the dataset curators via the provided email. }

\section*{Supplementary Materials (Long Code Arena)}

These supplementary materials include the following:

\begin{enumerate}
    \item \Cref{sec:sheet1} --- datasheet for the Library-based code generation dataset.
    \item \Cref{sec:sheet2} --- datasheet for the CI builds repair dataset.
    \item \Cref{sec:sheet3} --- datasheet for the Project-level code completion dataset.
    \item \Cref{sec:sheet4} --- datasheet for the Commit message generation dataset.
    \item \Cref{sec:sheet5} --- datasheet for the Bug localization dataset.
    \item \Cref{sec:sheet6} --- datasheet for the Module summarization dataset.
\end{enumerate}

\section{Datasheet for the Library-Based Code Generation Dataset}
\label{sec:sheet1}

\subsection{Motivation}


\begin{enumerate}[start=1,label={Q\arabic*}]
    \item \textbf{For what purpose was the dataset created?}
    \begin{itemize}
        \item The dataset for the library-based code generation task is a part of Long Code Arena, a set of six benchmarks that cover different aspects of code processing. The most important feature of Long Code Arena is utilization of module- or project-level contexts for all the tasks, code generation included. Thus, the purpose of this dataset is to evaluate how good machine learning models can utilize data from an entire software project when solving the code generation task.
    \end{itemize}
    
    \item \textbf{Who created this dataset (\textit{e.g.}, which team, research group) and on behalf of which entity (\textit{e.g.,} company, institution, organization)?}
    \begin{itemize}
        \item \qII
    \end{itemize}

    \item \textbf{Who funded the creation of the dataset?}
    \begin{itemize}
        \item \qIII
    \end{itemize}

    \item \textbf{Any other comments?}
    \begin{itemize}
        \item No.
    \end{itemize}
\end{enumerate}

\subsection{Composition}

\begin{enumerate}[start=5,label={Q\arabic*}]

    \item \textbf{What do the instances that comprise the dataset represent?}

    \begin{itemize}
        \item Each of the 150 samples in the dataset represents an instruction that a machine learning model should follow when generating a Python program, reference data for evaluation of the generation quality, and relevant data that can be used to improve generation. This relevant data is the source code of an entire Python library, based on a usage example from which we created the instruction for generation.
    \end{itemize}

    \item \textbf{How many instances are there in total (of each type, if appropriate)?}
    \begin{itemize}
        \item There are 150 datapoints in total.
    \end{itemize}

    \item \textbf{Does the dataset contain all possible instances or is it a sample (not necessarily random) of instances from a larger set?}
    \begin{itemize}
        \item \qVII It comes from a larger set of Python repositories.
    \end{itemize}

    \item \textbf{What data does each instance consist of?}
    \begin{itemize}
        \item The structure of the datapoints is presented in \Cref{tab:code_generation_dp_struct}.

\begin{table}[t]
\centering
\caption{The structure of datapoints in the library-based code generation dataset.}
\label{tab:code_generation_dp_struct}
\begin{tabular}{cp{7cm}}
\toprule
\textbf{Field} & \multicolumn{1}{c}{\textbf{Description}} \\
\midrule
\rowcolor{gray!30}\textbf{repo\_full\_name} & Concatenated repository name and owner \\
\textbf{repo\_name} & Library repository name \\
\rowcolor{gray!30}\textbf{repo\_owner} & Library repository owner \\
\textbf{instruction} & Task for code generation \\
\rowcolor{gray!30}\textbf{reference} & Reference program written by the library authors \\
\textbf{clean\_reference} & Reference program with comments removed \\
\rowcolor{gray!30}\textbf{path\_to\_reference\_file} & Path to the reference in the repository (removed in repository snapshots to prevent data leakages) \\
\textbf{path\_to\_examples\_folder} & Path to the directory with examples in the repository (removed in repository snapshots to prevent data leakages) \\
\rowcolor{gray!30}\textbf{n\_unique\_apis} & Number of calls to library-specific APIs in the reference program \\
\textbf{unique\_apis} & List of calls to library-specific APIs in the reference program \\
\rowcolor{gray!30}\textbf{project\_defined\_elements} & All class and method names in the repository \\
\textbf{api\_calls} & All API calls in the reference program \\
\rowcolor{gray!30}\textbf{internal\_apis} & All API calls to the respective library in the reference program  \\
\bottomrule
\end{tabular}
\end{table}

    \end{itemize}

    \item \textbf{Is there a label or target associated with each instance?}
    \begin{itemize}
        \item The labels are available in two forms: the reference program that was written by library authors as an example of library usage, and the list of library-specific API calls that the reference program makes. Both the program itself and the list of API calls can be used to assess the quality of a program generated by a machine learning model under evaluation.
    \end{itemize}

    \item \textbf{Is any information missing from individual instances?}
    \begin{itemize}
        \item No.
    \end{itemize}

    \item \textbf{Are relationships between individual instances made explicit?}
    \begin{itemize}
        \item \qXI
    \end{itemize}

    \item \textbf{Are there recommended data splits (\textit{e.g.}, training, development/validation, testing)?}
    \begin{itemize}
        \item \qXII
    \end{itemize}

    \item \textbf{Are there any errors, sources of noise, or redundancies in the dataset?}
    \begin{itemize}
        \item See the description of preprocessing in \ref{q_A_22}.
    \end{itemize}

    \item \textbf{Is the dataset self-contained, or does it link to or otherwise rely on external resources?}
    \begin{itemize}
        \item The dataset is self-contained, as it provides the snapshots of all associated repositories.
    \end{itemize}

    \item \textbf{
    Does the dataset contain data that might be considered confidential?
    }
    \begin{itemize}
        \item \qXV
    \end{itemize}

    \item \textbf{Does the dataset contain data that, if viewed directly, might be offensive, insulting, threatening, or might otherwise cause anxiety?}
    \begin{itemize}
        \item \qXVI
    \end{itemize}

    \item \textbf{Does the dataset relate to people?}

    \begin{itemize}
        \item \qXVII
    \end{itemize}

    \item \textbf{Does the dataset identify any subpopulations (\textit{e.g.}, by age, gender)?}
    \begin{itemize}
        \item \qXVIII
    \end{itemize}

    \item \textbf{Is it possible to identify individuals (\textit{i.e.}, one or more natural persons), either directly or indirectly (\textit{i.e.}, in combination with other data) from the dataset?}
    \begin{itemize}
        \item \qXIX
    \end{itemize}

    \item \textbf{Does the dataset contain data that might be considered sensitive in any way?}
    \begin{itemize}
        \item \qXX
    \end{itemize}

    \item \textbf{Any other comments?} 
    \begin{itemize}
        \item No.
    \end{itemize}
\end{enumerate}

\subsection{Collection}

\begin{enumerate}[start=22,label={Q\arabic*}]
    \item \textbf{How was the data associated with each instance acquired?}
    \label{q_A_22}
    \begin{itemize}
        \item To collect the data, we use the following protocol:

\begin{enumerate}
\item We collect repositories from GitHub with at least 1,000 commits, at least ten contributors, issues, and stars, at least 10,000 lines of code, not a fork, last commit after 01.06.2023, and a permissive license (we use the most popular permissive licenses --- MIT, Apache-2.0, BSD-3-Clause, and BSD-2-Clause). For the library-specific code generation task, we leave only repositories having Python as the main language.

\item For each repository, we detect the folder with usage examples: a folder with ``.py'' files that contains ``examples'' in its name. If a repository does not have such a folder, we filter it out. After this step, we are left with 883 repositories that have usage examples.

\item We then identify library-specific APIs for each of the 883 repositories. We extract all names of all methods, classes, and constants defined in these repositories, and treat as ``library-specific'' the ones that appear only in a single repository.

\item We then collect all Python files from the folders with examples and filter them: (i) remove examples shorter than 100 or longer than 40,000 characters (excluding comments), (ii) remove examples that have fewer than 400 characters of comments in order to then write high-quality instruction for generation, (iii) remove examples that use fewer than ten API calls specific to the given library. These filters result in 150 files (usage examples) from 62 libraries, with each file heavily relying on the APIs of the respective project.

\item After we have the usage examples for libraries, we create instructions for generating them. We first run the selected 150 files through GPT-4~\cite{gpt4}, prompting it to generate an instruction for generating the respective file. You can see the prompt for generation in \Cref{fig:code-generation-prompt}. This leaves us with step-by-step instructions that the LLM should then follow to generate a script that utilizes the library at hand. Then, we manually fix each instruction in order to reduce hinting to specific library methods and ensure their correctness. 
\end{enumerate}

\begin{figure}[h]
\centering
\begin{BVerbatim}
SYSTEM: We are developing a benchmark to assess quality of
code generation models. As a part of the benchmark, we include
the task of generating code based that uses the particular
library from a description in natural language. As a source of
data for this task we will use coding examples in Python
provided by library developers. Your task will be to generate
a text description of the provided Python code that will then
be used as an input for the generation task.

USER: Here is the code. You should write an instruction that
summarizes its contents and would allow another model to
generate this snippet of code, excluding the comments. Make
the instruction abstract, do not mention specific code
constructions that the generator should use. Be concise.
Generator will be able to access the contents of the following
library: [LIBRARY_NAME]. Use wording such as "Generate code
that ..." in your instruction.

[CODE]
\end{BVerbatim}
\caption{Prompt for generating instructions from library usage examples.}
\label{fig:code-generation-prompt}
\end{figure}

    \end{itemize}

    \item \textbf{What mechanisms or procedures were used to collect the data (\textit{e.g.}, hardware apparatus or sensor, manual human curation, software program, software API)?}
    \begin{itemize}
        \item We use GitHub Search~\cite{dabic2021data} to collect the initial list of repositories. We use GitHub API for data collection. We use OpenAI's GPT-4~\cite{gpt4} to generate instructions for code generation and then conduct manual curation of the instructions by the paper authors having more than six years of experience of software development in Python.
    \end{itemize}
    
    \item \textbf{
    If the dataset is a sample from a larger set, what was the sampling strategy?}
    \begin{itemize}
        \item The dataset is sampled from a larger set of repositories by selecting only repositories with Python as the main language and further filtering as described in \ref{q_A_22}.
    \end{itemize}

    \item \textbf{Who was involved in the data collection process (\textit{e.g.}, students, crowdworkers, contractors) and how were they compensated?}
    \begin{itemize}
        \item \qXXV
    \end{itemize}

    \item \textbf{Over what timeframe was the data collected?}
    \begin{itemize}
        \item The construction of this dataset took place between October 2023 and January 2024.
    \end{itemize}

    \item \textbf{Were any ethical review processes conducted?}
    \begin{itemize}
        \item No.
    \end{itemize}

    \item \textbf{Does the dataset relate to people?}
    \begin{itemize}
        \item \qXXVIII
    \end{itemize}

    \item \textbf{Did you collect the data from the individuals in question directly, or obtain it via third parties or other sources (\textit{e.g.}, websites)?}
    \begin{itemize}
        \item \qXXIX
    \end{itemize}

    \item \textbf{Were the individuals in question notified about the data collection?}
    \begin{itemize}
        \item \qXXX
    \end{itemize}

    \item \textbf{Did the individuals in question consent to the collection and use of their data?}
    \begin{itemize}
        \item \qXXXI
    \end{itemize}

    \item \textbf{If consent was obtained, were the consenting individuals provided with a mechanism to revoke their consent in the future or for certain uses?}
    \begin{itemize}
        \item \qXXXII
    \end{itemize}

    \item \textbf{Has an analysis of the potential impact of the dataset and its use on data subjects been conducted?}
    \begin{itemize}
        \item \qXXXIII
    \end{itemize}

    \item \textbf{Any other comments?} 
    \begin{itemize}
        \item No.
    \end{itemize}
\end{enumerate}

\subsection{Preprocessing / Cleaning / Labeling}

\begin{enumerate}[start=35,label={Q\arabic*}]
    \item \textbf{Was any preprocessing/cleaning/labeling of the data done?}
    \begin{itemize}
        \item We describe the steps for creating the dataset for library-specific code generation in \ref{q_A_22}.
    \end{itemize}

    \item \textbf{Was the “raw” data saved in addition to the preprocessed/cleaned/labeled data?}
    \begin{itemize}
        \item We include into the dataset both repository snapshots and human-written programs that served as a basis for the tasks. The larger set of repositories before filtering steps is not provided in the dataset. 
    \end{itemize}

    \item \textbf{Is the software used to preprocess/clean/label the instances available?}
    \begin{itemize}
        \item \qXXXVII
    \end{itemize}

    \item \textbf{Any other comments?} 
    \begin{itemize}
        \item No.
    \end{itemize}
\end{enumerate}

\subsection{Uses}\label{sec:supp-codegen-results}

\begin{enumerate}[start=39,label={Q\arabic*}]

    \item \textbf{Has the dataset been used for any tasks already?}
    \label{q_A_39}
    \begin{itemize}
        \item We use the dataset to assess the quality of models in the library-based code generation task. To do so, we develop and evaluate multiple baselines solutions, and propose two metrics for assessing quality:
        \begin{enumerate}
            \item We measure ChrF~\cite{chrf} between the generated code and the reference program written by developers of the respective library as a usage example. ChrF estimates similarity between two texts, or code snippets as in our case, using character n-grams. Previous study~\cite{evtikhiev2023out} has shown that it is more robust compared to other metrics when assessing code generation quality.
            \item We also propose to use \emph{API Recall}, the ratio of library-specific methods and classes used in the reference solution that also appear in the generated code. The metric assumes that a model that solves the generation task well should be able to identify useful APIs in the library, the same as library developers utilized in the provided usage example.
        \end{enumerate}
        
        \item We develop and evaluate baselines based on six popular LLMs in two setups. For models, we use proprietary GPT-3.5-turbo and GPT-4~\cite{gpt4}, and instruction-tuned versions of open-source CodeLlama-7B, CodeLlama-70B~\cite{codellama}, Mistral-7B~\cite{mistral}, and Mixtral-8x7B~\cite{mixtral}. In the first setup, we run the models without any information from the library aside from the instruction for generation that recommends using it. In the second setup, we treat instruction as a query and use BM-25 to find top-20 most relevant class and method names in the library. To do so, we also split the names by snake\_case and camelCase, remove punctuation from them, and turn them into lower case. Then, we add the retrieved method names to the instruction and propose the model under evaluation to use them.

        \item \Cref{tab:code_generation_results} shows the results of evaluation for the baselines. GPT-4 shows the best quality according to both metrics, with GPT-3.5 following it. Notably, CodeLlama-70B shows the worst quality by far. This happens because the model refused to answer the code generation request for most of the queries, answering with a stub message. Models aside from GPT-4 get very low API Recall, showing that they are not well familiar with the libraries that we want them to use. We treat this as a success for the benchmark, as it suggests that using open-source libraries (that models may have seen during training) does not make the task easy. Using a simplistic retrieval approach to enhance context allows to add a few points to API Recall for most models, however, the task remains far from being solved.

\begin{table}[t]

\caption{Results of baselines for the library-based code generation task.}
\label{tab:code_generation_results}
\centering
\begin{tabular}{lcccc}
\toprule
              & \multicolumn{2}{c}{\textbf{No context}} & \multicolumn{2}{c}{\textbf{With retrieved APIs}} \\\cmidrule(lr){2-3}\cmidrule(lr){4-5}
              & \textbf{ChrF}           & \textbf{API Recall}    & \textbf{ChrF}               & \textbf{API Recall}         \\ \midrule
GPT-4         & \textbf{0.41}  & \textbf{0.37} & \textbf{0.39}      & \textbf{0.36}      \\
GPT-3.5       & 0.26           & 0.17          & 0.26               & 0.19               \\
CodeLlama-7B  & 0.28           & 0.09          & 0.29               & 0.15               \\
Mistral-7B    & 0.30           & 0.07          & 0.31               & 0.13               \\
Mixtral-8x7B  & 0.29           & 0.11          & 0.29               & 0.13               \\
CodeLlama-70B & 0.06           & 0.02          & 0.11               & 0.04  \\    \bottomrule        
\end{tabular}
\end{table}

    \end{itemize}

        \item \textbf{Is there a repository that links to any or all papers or systems that use the dataset?}
    \begin{itemize}
        \item \qXL
    \end{itemize}
    
    \item \textbf{What tasks could the dataset be used for?}
    \begin{itemize}
        \item The dataset can be used for assessing models solving the library-based code generation task, as explained in \ref{q_A_39}.
    \end{itemize}
    
    \item \textbf{Is there anything about the composition of the dataset or the way it was collected and preprocessed/cleaned/labeled that might impact future uses?}
    \begin{itemize}
        \item \qXLII
    \end{itemize}
    
    \item \textbf{Are there tasks for which the dataset should not be used?}
    \begin{itemize}
        \item \qXLIII
    \end{itemize}

    \item \textbf{Any other comments?}
    \begin{itemize}
        \item No.
    \end{itemize}
\end{enumerate}

\subsection{Distribution}

\begin{enumerate}[start=45,label={Q\arabic*}]
    \item \textbf{Will the dataset be distributed to third parties outside of the entity?}
    \begin{itemize}
        \item \qXLV
    \end{itemize}

    \item \textbf{How will the dataset be distributed? Does the dataset have a digital object identifier (DOI)?}
    \begin{itemize}
        \item The dataset is available through DOI at the HuggingFace Hub: {\url{https://doi.org/10.57967/hf/2510}}.
    \end{itemize}

    \item \textbf{When will the dataset be distributed?}
    \begin{itemize}
        \item \qXLVII
    \end{itemize}

    \item \textbf{Will the dataset be distributed under a copyright or other intellectual property (IP) license, and/or under applicable terms of use (ToU)?}
    \begin{itemize}
        \item \qXLVIII
    \end{itemize}

    \item \textbf{Have any third parties imposed IP-based or other restrictions on the data associated with the instances?}
    \begin{itemize}
        \item No.
    \end{itemize}

    \item \textbf{Do any export controls or other regulatory restrictions apply to the dataset or to individual instances?}
    \begin{itemize}
        \item \qL
    \end{itemize}

    \item \textbf{Any other comments?}
    \begin{itemize}
        \item No.
    \end{itemize}
\end{enumerate}

\subsection{Maintenance}
\begin{enumerate}[start=52,label={Q\arabic*}]
    \item \textbf{Who is supporting/hosting/maintaining the dataset?}
    \begin{itemize}
        \item \qLII
    \end{itemize}

    \item \textbf{How can the owner/curator/manager of the dataset be contacted (\textit{e.g.}, email address)?}
    \begin{itemize}
        \item \qLIII
    \end{itemize}

    \item \textbf{Is there an erratum?}
    \begin{itemize}
        \item \qLIV
    \end{itemize}

    \item \textbf{Will the dataset be updated? (\textit{e.g.}, to correct labeling errors, add new instances, delete instances)?}
    \begin{itemize}
        \item \qLV
    \end{itemize}

    \item \textbf{If the dataset relates to people, are there applicable limits on the retention of the data associated with the instances?}
    \begin{itemize}
        \item \qLVI
    \end{itemize}

    \item \textbf{Will older versions of the dataset continue to be supported/hosted/maintained?}
    \begin{itemize}
        \item \qLVII
    \end{itemize}

    \item \textbf{If others want to extend/augment/build on/contribute to the dataset, is there a mechanism for them to do so?}
    \begin{itemize}
        \item \qLVIII
    \end{itemize}

    \item \textbf{Any other comments?} 
    \begin{itemize}
        \item No.
    \end{itemize}
\end{enumerate}

\section{Datasheet for the CI Builds Repair dataset}
\label{sec:sheet2}

\subsection{Motivation}


\begin{enumerate}[start=1,label={Q\arabic*}]
    \item \textbf{For what purpose was the dataset created?}
    \begin{itemize}
        \item CI builds repair dataset is a part of the Long Code Arena aimed at evaluating models on repository-level long-context real-life tasks. CI builds repair benchmark is aimed at testing models in fixing real-life issues in continuous integration. We use the functionality and data of GitHub Actions~\cite{actions}, a popular continuous integration and continuous deployment (CI/CD) service. The minimal set of data for the task consists of a repository snapshot at the commit that caused the failure of the CI workflow and the logs of the failed step. Based on the provided data, the model under evaluation has to generate a patch for the project that will make the build pass. The testing then happens by running CI workflows for the repository with the generated patch.
    \end{itemize}
    
    \item \textbf{Who created this dataset (\textit{e.g.}, which team, research group) and on behalf of which entity (\textit{e.g.,} company, institution, organization)?}
    \begin{itemize}
        \item \qII
    \end{itemize}

    \item \textbf{Who funded the creation of the dataset?}
    \begin{itemize}
        \item \qIII
    \end{itemize}

    \item \textbf{Any other comments?}
    \begin{itemize}
        \item No.
    \end{itemize}
\end{enumerate}

\subsection{Composition}

\begin{enumerate}[start=5,label={Q\arabic*}]

    \item \textbf{What do the instances that comprise the dataset represent?}

    \begin{itemize}
        \item The dataset instances for the CI builds repair task consist of a repository snapshot at the commit with failing CI, the logs of the failed CI step, a diff that fixes the CI, and various metadata. We include diffs to help dataset users to compare the answers of their models with a ground truth solution. We do not store repository snapshots and fetch them from GitHub during benchmarking to reduce the dataset's memory requirements. To ensure the repositories are available, we forked them to a separate organization.
    \end{itemize}

    \item \textbf{How many instances are there in total (of each type, if appropriate)?}
    \begin{itemize}
        \item There are 77 datapoints in total.
    \end{itemize}

    \item \textbf{Does the dataset contain all possible instances or is it a sample (not necessarily random) of instances from a larger set?}
    \begin{itemize}
        \item \qVII It comes from a larger set of GitHub Actions builds in Python repositories.
    \end{itemize}

    \item \textbf{What data does each instance consist of?}
    \begin{itemize}
        \item The structure of the datapoints is presented in \Cref{tab:dp_struct}.

\begin{table}[t]
\centering
\caption{The structure of datapoints in the CI builds repair dataset.}
\label{tab:dp_struct}
\begin{tabular}{cp{8cm}}
\toprule
\textbf{Field} & \multicolumn{1}{c}{\textbf{Description}} \\
\midrule
\rowcolor{gray!30}\textbf{contributor} & The username of the contributor that committed changes \\
\textbf{difficulty} & The difficulty of the problem according to an assessor on a 1--3 scale \\
\rowcolor{gray!30}\textbf{diff} & Contents of the diff between the failed and the successful commits \\
\textbf{head\_branch} & Name of the original branch that the commit was pushed to \\
\rowcolor{gray!30}\textbf{id} & Unique ID of the datapoint \\
\textbf{language} & The main language of the repository \\
\rowcolor{gray!30}\textbf{logs} & List of dictionaries with logs of the failed job and name of the failed step in this job\\
\textbf{repo\_name} & Name of the original repository \\
\rowcolor{gray!30}\textbf{repo\_owner} & Owner of the original repository \\
\textbf{sha\_fail} & SHA of the failed commit \\
\rowcolor{gray!30}\textbf{sha\_success} & SHA of the successful commit \\
\textbf{workflow} & Contents of the workflow file \\
\rowcolor{gray!30}\textbf{workflow\_filename} & The name of the workflow file (without full path) \\
\textbf{workflow\_name} & The name of the workflow \\
\rowcolor{gray!30}\textbf{workflow\_path} & The full path to the workflow file \\
\textbf{changed\_files} & List of files changed in the diff \\
\rowcolor{gray!30}\textbf{commit\_link} & URL to a commit corresponding to the failed job \\
\bottomrule

\end{tabular}

\end{table}

    \end{itemize}

    \item \textbf{Is there a label or target associated with each instance?}
    \begin{itemize}
        \item There is no label or target in the dataset. The goal of the benchmark is to submit a fix to a GitHub repository that will make the CI build pass. We provide code for evaluation in our GitHub repository.
    \end{itemize}

    \item \textbf{Is any information missing from individual instances?}
    \begin{itemize}
        \item No.
    \end{itemize}

    \item \textbf{Are relationships between individual instances made explicit?}
    \begin{itemize}
        \item \qXI
    \end{itemize}

    \item \textbf{Are there recommended data splits (\textit{e.g.}, training, development/validation, testing)?}
    \begin{itemize}
        \item \qXII
    \end{itemize}

    \item \textbf{Are there any errors, sources of noise, or redundancies in the dataset?}
    \begin{itemize}
        \item We describe the preprocessing strategy in~\ref{ci-gather-protocol} and discuss the possible obsoletion of the datapoints in~\ref{ci-future-errosion}.
    \end{itemize}

    \item \textbf{Is the dataset self-contained, or does it link to or otherwise rely on external resources?}
    \begin{itemize}
        \item The dataset does not store the repository snapshots but rather fetches them from GitHub during benchmarking to reduce the dataset’s memory requirements. Otherwise, the dataset is self-contained.
    \end{itemize}

    \item \textbf{
    Does the dataset contain data that might be considered confidential?
    }
    \begin{itemize}
        \item \qXV
    \end{itemize}

    \item \textbf{Does the dataset contain data that, if viewed directly, might be offensive, insulting, threatening, or might otherwise cause anxiety?}
    \begin{itemize}
        \item \qXVI
    \end{itemize}

    \item \textbf{Does the dataset relate to people?}

    \begin{itemize}
        \item \qXVII
    \end{itemize}

    \item \textbf{Does the dataset identify any subpopulations (\textit{e.g.}, by age, gender)?}
    \begin{itemize}
        \item \qXVIII
    \end{itemize}

    \item \textbf{Is it possible to identify individuals (\textit{i.e.}, one or more natural persons), either directly or indirectly (\textit{i.e.}, in combination with other data) from the dataset?}
    \begin{itemize}
        \item \qXIX
    \end{itemize}

    \item \textbf{Does the dataset contain data that might be considered sensitive in any way?}
    \begin{itemize}
        \item \qXX
    \end{itemize}

    \item \textbf{Any other comments?} 
    \begin{itemize}
        \item No.
    \end{itemize}
\end{enumerate}

\subsection{Collection}

\begin{enumerate}[start=22,label={Q\arabic*}]
    \item \textbf{How was the data associated with each instance acquired?}\label{ci-gather-protocol}
    \begin{itemize}
        \item To collect the data, we used the following protocol:

\begin{enumerate}
\item For all the collected Python repositories, we get the full list of the actions run in the repository, limited to last 90 days. Downloaded data contains action status (failed or successful) and links to the action runs.

\item We gather a list of pairs of consecutive commits in which the first commit causes a failure of a workflow but the next one makes it build successfully.

\item For each pair of commits, we download:

\begin{itemize}
\item logs of the failed step of the failed commit;
\item diff between the failed and successful commit (\textit{correction diff});
\item metadata of the failed commit.
\end{itemize}

During the download, we clean the data according to the following filters (on the fly, to avoid excessive requests to GitHub API):

\begin{itemize}
\item To reduce the benchmarking time, we eliminate runs that take more than 10 minutes (measured on successful action run).

\item To minimize the number of actions that contain pure formatting issues, we filter out datapoints, in which the names of the workflow, target, or failed step contain any of the following substrings: \{\textit{mypy}, \textit{lint}, \textit{flake8}, \textit{black}\}. We allow these substrings in the target name if there is more than one target in the action run.

\item We remove runs for which the workflow file contains substrings \{\textit{token}, \textit{secret}\} to ensure that we can run them without any prerequisites.

\item We keep only datapoints for which the correction diff (i) contains at least one \texttt{.py} file, and (ii) only contains files that match either of the following items: \{code file, \textit{*.md}, \textit{*.rst}, \textit{LICENSE*}, \textit{readme*}, \textit{doc/*}\}. We do so to ensure that there are no changes in artifacts such as resources or data files, which the model cannot fix given the present context.

\end{itemize}

\item To isolate the problem to a single issue per datapoint, when running the benchmark, we delete all \texttt{.yaml} files in the \texttt{.github/workflows/} directory, ensuring that only this workflow would be run. We also remove workflows that contain links to other workflow files to make sure that the target workflow is independent.

\end{enumerate}
    \end{itemize}

    \item \textbf{What mechanisms or procedures were used to collect the data (\textit{e.g.}, hardware apparatus or sensor, manual human curation, software program, software API)?}
    \begin{itemize}
        \item We use GitHub API to collect the data and further manual verification and assessment to filter it.
    \end{itemize}
    
    \item \textbf{
    If the dataset is a sample from a larger set, what was the sampling strategy?}
    \begin{itemize}
        \item The dataset is sampled from a larger set of repositories by selecting only repositories with Python as the main language. Also, we only collect CI builds over the period of 90 days and then filter them as described in \ref{ci-gather-protocol}.
    \end{itemize}

    \item \textbf{Who was involved in the data collection process (\textit{e.g.}, students, crowdworkers, contractors) and how were they compensated?}
    \begin{itemize}
        \item \qXXV
    \end{itemize}

    \item \textbf{Over what timeframe was the data collected?}
    \begin{itemize}
        \item The dataset has been collected in December of 2023. Only datapoints spanning three months before collection have been gathered, since logs of the GitHub Actions are stored only for 90 days. Thus, the dataset collection timeframe is October--December of 2023. 
    \end{itemize}

    \item \textbf{Were any ethical review processes conducted?}
    \begin{itemize}
        \item No.
    \end{itemize}

    \item \textbf{Does the dataset relate to people?}
    \begin{itemize}
        \item \qXXVIII
    \end{itemize}

    \item \textbf{Did you collect the data from the individuals in question directly, or obtain it via third parties or other sources (\textit{e.g.}, websites)?}
    \begin{itemize}
        \item \qXXIX
    \end{itemize}

    \item \textbf{Were the individuals in question notified about the data collection?}
    \begin{itemize}
        \item \qXXX
    \end{itemize}

    \item \textbf{Did the individuals in question consent to the collection and use of their data?}
    \begin{itemize}
        \item \qXXXI
    \end{itemize}

    \item \textbf{If consent was obtained, were the consenting individuals provided with a mechanism to revoke their consent in the future or for certain uses?}
    \begin{itemize}
        \item \qXXXII
    \end{itemize}

    \item \textbf{Has an analysis of the potential impact of the dataset and its use on data subjects been conducted?}
    \begin{itemize}
        \item \qXXXIII
    \end{itemize}

    \item \textbf{Any other comments?} 
    \begin{itemize}
        \item No.
    \end{itemize}
\end{enumerate}

\subsection{Preprocessing / Cleaning / Labeling}

\begin{enumerate}[start=35,label={Q\arabic*}]
    \item \textbf{Was any preprocessing/cleaning/labeling of the data done?}
    \begin{itemize}
        \item The basic data filters are described in the data collection procedure in \ref{ci-gather-protocol}. Here, we provide further filtering steps.
\begin{enumerate}
\item We limited ourselves to the 100 largest Python repositories (main language: Python, the ratio of the main language $>$ 0.95) with permissive licences. From each repository, we take no more than three branches, for each branch --- no more than three different workflows, and for each workflow --- no more than three datapoints. Thus, each repository could contribute up to 27 datapoints. The automated data collection process resulted in 336 datapoints (see Table~\ref{tab:data_num}).

\item The human assessor assessed the datapoints to verify that logs contain all the necessary information to fix the issue and graded the datapoints on a 1--3 scale according to their difficulty. \Cref{tab:diffic} describes the difficulty levels and the sizes of the available buckets. 

\item In the last step, we run all datapoints through our benchmark at both the failed and the successful commit. We then keep only the datapoints that remained failing / passing at the respective commits. Moreover, we repeat the procedure after 6 months from the initial procedure to ensure the durability of the dataset. This last step is crucial as it filtered out ~50\% of the datapoints: quite many passing workflows started failing due to changes in library versions that were not specified by repository owners, connection issues, missing remote files or certificates. \Cref{tab:data_num} reports the number of filtered datapoints at each step.
\end{enumerate}

\begin{table}[t]
    \centering
\caption{Number of datapoints on each mining step.}
\label{tab:data_num}
    \begin{tabular}{cc} \toprule 
         \textbf{Data mining step} & \textbf{\# of datapoints}
\\ \midrule
         \rowcolor{gray!30}Initial set of sampled workflows & 336
\\ 
         Datapoints that passed assessor verification & 210
\\ 
         \rowcolor{gray!30}Datapoints that passed GitHub Actions & 144
\\ 
         Datapoints that passed GitHub Actions after 6 months & 77
\\ \bottomrule
    \end{tabular}

\end{table}

\begin{table}[t]
\centering
\caption{Data split by the difficulty.}
\label{tab:diffic}
\begin{tabular}{ccp{5.5cm}}
\toprule
\textbf{Difficulty} & \textbf{\# of datapoints} & \multicolumn{1}{c}{\textbf{Description}} \\
\midrule
\rowcolor{gray!30}1 & 35 & Issues with formatting \\
2 & 14 & Local issues or issues with typing \\
\rowcolor{gray!30}3 & 28 & Issues that require information about other files in the repository \\
\midrule
\textbf{Total} & \textbf{77} &   \\
\bottomrule

\end{tabular}

\end{table}

    \end{itemize}

    \item \textbf{Was the “raw” data saved in addition to the preprocessed/cleaned/labeled data?}
    \begin{itemize}
        \item No.
    \end{itemize}

    \item \textbf{Is the software used to preprocess/clean/label the instances available?}
    \begin{itemize}
        \item \qXXXVII
    \end{itemize}

    \item \textbf{Any other comments?} 
    \begin{itemize}
        \item  Context-related statistics are presented in Table~\ref{tab:context_stats}.

\begin{table}[t]
    \centering
\caption{Context-related statistics.}
\label{tab:context_stats}
    \begin{tabular}{ccc} \toprule 
         \textbf{Context metric} & \textbf{Mean} & \textbf{Median}
\\ \midrule
         Symbols in logs & 145K & 6.5K
\\ 
         Files in repository & 610 & 240
\\ 
         Lines in repository & 170K & 56K
\\ 
         Symbols in repository & 7.5M & 2.4M
\\ \bottomrule
    \end{tabular}

\end{table}
        
    \end{itemize}
\end{enumerate}

\subsection{Uses}

\begin{enumerate}[start=39,label={Q\arabic*}]

    \item \textbf{Has the dataset been used for any tasks already?}
    \begin{itemize}
        \item We use the collected dataset to assess multiple LLMs in the CI builds repair task. 
        To make the task easier to tackle, we provide models with an oracle --- when asking to fix the build, we also provide the list of files and specific code blocks in them that should be fixed. The information on which files need fixing comes from the ground truth commit that fixed the build. In the future, if the task becomes too easy for the models, oracle can be simply removed to make the task even more realistic and challenging.

        \item To prompt the models to solve the task, we use the following strategy. To prepare an instruction, we locate the first occurrence of the case-insensitive substring ``error'' in the logs and take a 7-line context around this occurrence (3 lines before and after). If the substring is not found, we use 7 last log lines. The instruction then reads as follows: ``\textit{Fix CI in order for tests to pass. Relevant logs: \{relevant\_logs\}}''.
        We then prompt the LLM to modify the code blocks provided by an oracle to align with the given instructions, and pass all the code blocks in a single request in the following way:

\vspace{0.2cm}

\begin{BVerbatim}
[start of file.py#L12]
...code line 12...
...code line 13...
...
[end of file.py#L12]
\end{BVerbatim}
\vspace{0.2cm}

\item After an LLM replies with the edited versions of the code sections, we convert them into a diff and apply the resulting patch to the repository. Then, the developed benchmark sends the updated version of the repository to GitHub Actions and collects the results. \Cref{tab:CI-bench-res} shows the evaluation results for several models: proprietary GPT-3.5 and GPT-4~\cite{gpt4}, open-source versions of Llama-2~\cite{touvron2023llama}, Mistral-7B~\cite{mistral}, and Mixtral-8x7B~\cite{mixtral}.

\begin{table}[t]
    \centering
\caption{Pass@1 scores of the CI builds repair benchmark for various LLMs}
\label{tab:CI-bench-res}
    \begin{tabular}{cc} \toprule 
         \textbf{Model} & \textbf{Pass@1}
\\ \midrule 
        Mistral-7B & 0.065
\\ 
         Mixtral-8x7B & 0.039
\\  
         Llama-2-7B & 0.065
\\ 
         Llama-2-13B & 0.065
\\ 
         Llama-2-34B & 0.091
\\ 
         GPT-3.5 & \textbf{0.169}
\\ 
         GPT-4 & 0.156

\\ \bottomrule
    \end{tabular}

\end{table}
    \end{itemize}

    \item \textbf{Is there a repository that links to any or all papers or systems that use the dataset?}
    \begin{itemize}
        \item \qXL
    \end{itemize}

    \item \textbf{What tasks could the dataset be used for?}
    \begin{itemize}
        \item We implement the benchmark for using the CI builds repair dataset in our repository. The benchmark requires a user-implemented function (\textit{fix\_repo\_function}) that repairs locally stored repository, given the logs of a failing build. The procedure is the following:

\begin{enumerate}

\item The benchmark clones each repository snapshot with depth equal to 1 to a local machine.
\item Then, the benchmark runs the model under evaluation, which takes a datapoint  as input (mainly --- log and workflow files) and needs to repair the repository on the local machine by editing or replacing files.
\item The benchmark edits the workflow files to run only one workflow.
\item Then, it pushes the current state of the repository to a new branch in the separate GitHub organization.
\item When results of builds in GitHub Actions become available, the benchmark collects, analyzes, and returns them.

\end{enumerate}

\item To use the benchmark, one needs to send a request to join the GitHub organization\footnote{GitHub Organization for the benchmark: \url{https://github.com/LCA-CI-builds-repair}} since the procedure requires pushing changes to repositories in that organization. Moreover, keeping repositories as forks in a separate organization ensures that they will remain available.
The function \textit{fix\_repo\_function} takes the following (all optional) arguments:  
\begin{enumerate}

 \item \textbf{datapoint}:  datapoint from the dataset

 \item \textbf{repo\_path}:  path to the repository on the user's machine  

 \item \textbf{repo}:       git.Repo object from the GitPython library  

\item \textbf{out\_folder}: directory for outputting the benchmark results  
\end{enumerate}

\item Intermediate results contain datapoint ID and meta information, as well as the SHA of the commit pushed to the target repository. After collecting the results, the benchmark adds the status of the GitHub Actions build to this information. 
    \end{itemize}

    \item \textbf{Is there anything about the composition of the dataset or the way it was collected and preprocessed/cleaned/labeled that might impact future uses?}
    \begin{itemize}
        \item \qXLII
    \end{itemize}
    
    \item \textbf{Are there tasks for which the dataset should not be used?}
    \begin{itemize}
        \item \qXLIII
    \end{itemize}

    \item \textbf{Any other comments?}
    \begin{itemize}
        \item No.
    \end{itemize}
\end{enumerate}

\subsection{Distribution}

\begin{enumerate}[start=45,label={Q\arabic*}]
    \item \textbf{Will the dataset be distributed to third parties outside of the entity?}
    \begin{itemize}
        \item \qXLV
    \end{itemize}

    \item \textbf{How will the dataset be distributed? Does the dataset have a digital object identifier (DOI)?}
    \begin{itemize}
        \item The dataset is available through DOI at the HuggingFace Hub: {\url{https://doi.org/10.57967/hf/2511}}.
    \end{itemize}

    \item \textbf{When will the dataset be distributed?}
    \begin{itemize}
        \item \qXLVII. 
    \end{itemize}

    \item \textbf{Will the dataset be distributed under a copyright or other intellectual property (IP) license, and/or under applicable terms of use (ToU)?}
    \begin{itemize}
        \item \qXLVIII
    \end{itemize}

    \item \textbf{Have any third parties imposed IP-based or other restrictions on the data associated with the instances?}
    \begin{itemize}
        \item No.
    \end{itemize}

    \item \textbf{Do any export controls or other regulatory restrictions apply to the dataset or to individual instances?}
    \begin{itemize}
        \item \qL
    \end{itemize}

    \item \textbf{Any other comments?}
    \begin{itemize}
        \item No.
    \end{itemize}
\end{enumerate}

\subsection{Maintenance}
\begin{enumerate}[start=52,label={Q\arabic*}]
    \item \textbf{Who is supporting/hosting/maintaining the dataset?}
    \begin{itemize}
        \item \qLII
    \end{itemize}

    \item \textbf{How can the owner/curator/manager of the dataset be contacted (\textit{e.g.}, email address)?}
    \begin{itemize}
        \item \qLIII
    \end{itemize}

    \item \textbf{Is there an erratum?}
    \begin{itemize}
        \item \qLIV
    \end{itemize}

    \item \textbf{Will the dataset be updated? (\textit{e.g.}, to correct labeling errors, add new instances, delete instances)?}\label{ci-future-errosion}
    \begin{itemize}
        \item The dataset will be extended to more languages and samples in the future work.
        Also, since the task assessment relies on a loosely controlled GitHub Actions framework, there is a risk that some datapoints may become invalid over the course of time, as has already happened over the 6 months after the data gathering. We will continue updating the dataset with new datapoints and removing the ones that become obsolete with time.
    \end{itemize}

    \item \textbf{If the dataset relates to people, are there applicable limits on the retention of the data associated with the instances?}
    \begin{itemize}
        \item \qLVI
    \end{itemize}

    \item \textbf{Will older versions of the dataset continue to be supported/hosted/maintained?}
    \begin{itemize}
        \item \qLVII
    \end{itemize}

    \item \textbf{If others want to extend/augment/build on/contribute to the dataset, is there a mechanism for them to do so?}
    \begin{itemize}
        \item \qLVIII
    \end{itemize}

    \item \textbf{Any other comments?} 
    \begin{itemize}
        \item No.
    \end{itemize}
\end{enumerate}

\section{Datasheet for the Project-Level Code Completion Dataset}
\label{sec:sheet3}

\subsection{Motivation}


\begin{enumerate}[start=1,label={Q\arabic*}]
    \item \textbf{For what purpose was the dataset created?}
    \begin{itemize}
        \item Project-level code completion dataset is a part of Long Code Arena suite of benchmarks. The dataset can be used to evaluate approaches in utilizing long context in the code completion task.  In this dataset, we avoid possible data leakages by analyzing Git history, introduce a classification of completion lines, and provide entire repositories as a context. The benchmark is composed of four self-sufficient sets with various context sizes.

    \end{itemize}
    
    \item \textbf{Who created this dataset (\textit{e.g.}, which team, research group) and on behalf of which entity (\textit{e.g.,} company, institution, organization)?}
    \begin{itemize}
        \item \qII
    \end{itemize}

    \item \textbf{Who funded the creation of the dataset?}
    \begin{itemize}
        \item \qIII
    \end{itemize}

    \item \textbf{Any other comments?}
    \begin{itemize}
        \item No.
    \end{itemize}
\end{enumerate}

\subsection{Composition}

\begin{enumerate}[start=5,label={Q\arabic*}]

    \item \textbf{What do the instances that comprise the dataset represent?}

    \begin{itemize}
        \item Each instance that comprises the dataset consists of three key elements: a repository snapshot, a completion file, and target lines for the completion task. A repository snapshot is a list of all the filenames and contents of all text files from the repository (code, documentation, etc.). The state of the repository is before the commit where the completion file was added. A completion file is a Python file added in a particular commit. Target lines are a list of lines from the completion file that the model under evaluation should generate. Each line is also assigned one of classes that we describe in \ref{code-completion-classification}.
    \end{itemize}

    \item \textbf{How many instances are there in total (of each type, if appropriate)?}
    \begin{itemize}
        \item There are 934 datapoints in total, divided between four sets. Note that while each datapoint contains a single completion file, it requires the model to generate multiple lines in it.
        \begin{enumerate}
            \item \textit{small-context} set contains 144 datapoints.
            \item \textit{medium-context} set contains 224 datapoints.
            \item \textit{large-context} set contains 270 datapoints.
            \item \textit{huge-context} set contains 296 datapoints.
        \end{enumerate}
    \end{itemize}

    \item \textbf{Does the dataset contain all possible instances or is it a sample (not necessarily random) of instances from a larger set?}
    \begin{itemize}
        \item \qVII It comes from a larger set of Python repositories and commits.
    \end{itemize}

    \item \textbf{What data does each instance consist of?}
    \begin{itemize}
        \item The structure of datapoints:
        \begin{itemize}
            \item \texttt{repo} -- repository name in the format \texttt{\{GitHub\_user\_name\}\_\_\{repository\_name\}}
            \item \texttt{commit\_hash} -- hash of the commit where the completion file was added
            \item \texttt{completion\_file} -- dictionary with the completion file content in the following format:
            \begin{itemize}
                \item \texttt{filename} -- path to the completion file
                \item \texttt{content} -- content of the completion file 
            \end{itemize}
            \item \texttt{completion\_lines} -- dictionary where keys are categories of lines and values are a list of integers (numbers of lines to complete). The categories are described in \ref{code-completion-classification}.
            \item \texttt{repo\_snapshot} -- dictionary with a snapshot of the repository before the commit. Has the same structure as \texttt{completion\_file}, but filenames and contents are organized as lists.
            \item \texttt{completion\_lines\_raw} -- the same as \texttt{completion\_lines}, but before sampling.
        \end{itemize}
    \end{itemize}

    \item \textbf{Is there a label or target associated with each instance?}
    \begin{itemize}
        \item Targets for the completion task are provided in the \texttt{completion\_lines} field. To get a target line for completion, split the completion file by newline characters and select lines using the provided indices. Line categories are also provided.
    \end{itemize}

    \item \textbf{Is any information missing from individual instances?}
    \begin{itemize}
        \item No. However, during the collection process we focused only on the text-based files. While filenames for all files are included in the repository snapshot, the content of non-text files (\eg images) is recorded as \texttt{None}.
    \end{itemize}

    \item \textbf{Are relationships between individual instances made explicit?}
    \begin{itemize}
        \item \qXI
    \end{itemize}

    \item \textbf{Are there recommended data splits (\textit{e.g.}, training, development/validation, testing)?}
    \begin{itemize}
        \item \qXII
    \end{itemize}

    \item \textbf{Are there any errors, sources of noise, or redundancies in the dataset?}
    \begin{itemize}
        \item The repository snapshots are intentionally not filtered to ensure that all possible information could be utilized. As a result, the dataset includes sources of noise, such as auto-generated files, CSV data, etc.
    \end{itemize}

    \item \textbf{Is the dataset self-contained, or does it link to or otherwise rely on external resources?}
    \begin{itemize}
        \item The dataset is self-contained.
    \end{itemize}

    \item \textbf{
    Does the dataset contain data that might be considered confidential?
    }
    \begin{itemize}
        \item \qXV
    \end{itemize}

    \item \textbf{Does the dataset contain data that, if viewed directly, might be offensive, insulting, threatening, or might otherwise cause anxiety?}
    \begin{itemize}
        \item \qXVI
    \end{itemize}

    \item \textbf{Does the dataset relate to people?}

    \begin{itemize}
        \item \qXVII
    \end{itemize}

    \item \textbf{Does the dataset identify any subpopulations (\textit{e.g.}, by age, gender)?}
    \begin{itemize}
        \item \qXVIII
    \end{itemize}

    \item \textbf{Is it possible to identify individuals (\textit{i.e.}, one or more natural persons), either directly or indirectly (\textit{i.e.}, in combination with other data) from the dataset?}
    \begin{itemize}
        \item \qXIX
    \end{itemize}

    \item \textbf{Does the dataset contain data that might be considered sensitive in any way?}
    \begin{itemize}
        \item \qXX
    \end{itemize}

    \item \textbf{Any other comments?} 
    \begin{itemize}
        \item No.
    \end{itemize}
\end{enumerate}

\subsection{Collection}

\begin{enumerate}[start=22,label={Q\arabic*}]
    \item \textbf{How was the data associated with each instance acquired?}
    \begin{itemize}
        \item Starting with the common corpus of repositories, we then follow the following process to acquire the data:
        \begin{enumerate}
            \item \textbf{Traverse Git history}: We collect commits that add at least one new \texttt{.py} file. These files are candidates for the completion files.
            \item \textbf{Filtering collected commits}: We filter the commits to retain only those with the potential completion files containing between 200 and 2,000 lines, and with creation dates after January 1st, 2022.
            \item \textbf{Extract repository snapshots}: We create snapshots of the repositories based on the filtered commits, ensuring that we capture the state of the repository before the collected commit.
            \item \textbf{Split by the size of relevant context}: We split all the data into four groups based on the number of characters in \texttt{.py} files from the repository snapshots. The groups are: 
                (i) \textit{small-context}: less than $48K$ characters; 
                (ii) \textit{medium-context}: from $48K$ to $192K$ characters;
                (iii) \textit{large-context}: from $192K$ to $768K$ characters;
                (iv) \textit{huge-context}: more than $768K$ characters;
            \item \textbf{Sample datapoints}: we randomly sample 5 datapoints for each repository, and we randomly sample 75 repositories for each group. If fewer than 5 datapoints or 75 repositories are available, we use all available datapoints or repositories. We keep all 80 repositories for the \textit{medium-context} dataset.
            \item \textbf{Classify lines}: We perform line classification that is introduced in the paper and assign a main category to each line of the completion file.
            \item \textbf{Sample completion lines}: We sample lines from each category such that the average number of lines is no more than 5 for \textit{non-informative} and \textit{random} categories, and no more than 10 for other categories.
        \end{enumerate}
    \end{itemize}

    \item \textbf{What mechanisms or procedures were used to collect the data (\textit{e.g.}, hardware apparatus or sensor, manual human curation, software program, software API)?}
    \begin{itemize}
        \item Data collection utilized GitHub API. Further, we used manual verification and assessment for data filtering.
    \end{itemize}
    
    \item \textbf{
    If the dataset is a sample from a larger set, what was the sampling strategy?}
    \begin{itemize}
        \item We use the following sampling strategy for the datapoints when creating a dataset from a larger set of GitHub repositories:
            \begin{enumerate}
                \item If there are more than 5 datapoints from the same repository in a dataset, randomly sample 5.
                \item If there are more than 75 different repositories in a dataset, randomly sample 75. We keep all 80 repositories for the \textit{medium-context} set.
            \end{enumerate}
        \item We also filter the completion files:
            \begin{enumerate}
                \item The file contains from 200 to 2,000 lines.
                \item The file was added to the repository after January 1st, 2022.
            \end{enumerate}
        \item Finally, we sample the completion lines:
            \begin{enumerate}
                \item Sample 5 lines for \textit{non-informative} and \textit{random} categories.
                \item Remove exact duplicates by sampling 1 line from a set of exact duplicates.
                \item For each class except \textit{non-informative} and \textit{random}, remove 1 randomly chosen line from a datapoint with a maximum number of lines until we have an average not greater than 10. 
            \end{enumerate}
    \end{itemize}

    \item \textbf{Who was involved in the data collection process (\textit{e.g.}, students, crowdworkers, contractors) and how were they compensated?}
    \begin{itemize}
        \item \qXXV
    \end{itemize}

    \item \textbf{Over what timeframe was the data collected?}
    \begin{itemize}
        \item The dataset has been collected in December of 2023. Considering the filtering process, the data within the dataset spans from January 2022 to December 2023.
    \end{itemize}

    \item \textbf{Were any ethical review processes conducted?}
    \begin{itemize}
        \item No.
    \end{itemize}

    \item \textbf{Does the dataset relate to people?}
    \begin{itemize}
        \item \qXXVIII
    \end{itemize}

    \item \textbf{Did you collect the data from the individuals in question directly, or obtain it via third parties or other sources (\textit{e.g.}, websites)?}
    \begin{itemize}
        \item \qXXIX
    \end{itemize}

    \item \textbf{Were the individuals in question notified about the data collection?}
    \begin{itemize}
        \item \qXXX
    \end{itemize}

    \item \textbf{Did the individuals in question consent to the collection and use of their data?}
    \begin{itemize}
        \item \qXXXI
    \end{itemize}

    \item \textbf{If consent was obtained, were the consenting individuals provided with a mechanism to revoke their consent in the future or for certain uses?}
    \begin{itemize}
        \item \qXXXII
    \end{itemize}

    \item \textbf{Has an analysis of the potential impact of the dataset and its use on data subjects been conducted?}
    \begin{itemize}
        \item \qXXXIII
    \end{itemize}

    \item \textbf{Any other comments?} 
    \begin{itemize}
        \item No.
    \end{itemize}
\end{enumerate}

\subsection{Preprocessing / Cleaning / Labeling}

\begin{enumerate}[start=35,label={Q\arabic*}]
    \item \textbf{Was any preprocessing/cleaning/labeling of the data done?}\label{code-completion-classification}
    \begin{itemize}
        \item Classification of the lines is done for each of the completion files.
        There are six categories of completion lines according to various completion scenarios.
        \begin{enumerate}
            \item \textit{infile} -- a line contains at least one function or class that was declared in the completion file.
            \item \textit{inproject} -- a line contains at least one function or class that was declared in the repository snapshot files.
            \item \textit{common} -- a line contains at least one function or class that was classified to be common, \textit{e.g.}, \texttt{main}, \texttt{get}, etc.
            \item \textit{committed} -- a line contains at least one function or class that was declared in the files that were created in the same commit as the completion file (excluding the completion file).
            \item \textit{non-informative} -- a line that satisfies at least on of the following criteria: 
                (i) shorter than 5 characters or longer than 150 characters,
                (ii) a line with \texttt{print},
                (iii) a line with \texttt{import},
                (iv) a declaration of a function or a class,
                (v) a comment or contains an inline comment.
            \item \textit{random} -- all the lines that do not have any category.
        \end{enumerate}
        \item Some lines may have more than one category after the classification. We additionally identify the main category for each line based on the following approach.
        \begin{itemize}
            \item If a line has a \textit{committed} category, then the main category is \textit{committed}.
            \item If a line does not satisfy the previous condition, but has an \textit{inproject} category, then the main category is \textit{inproject}.
            \item If a line does not satisfy previous conditions, but has an \textit{infile} category, then the main category is \textit{infile}.
            \item If a line does not satisfy previous conditions, but has a \textit{common} category, then the main category is \textit{common}.
            \item If a line has a \textit{non-informative} category, then the main category is \textit{non-informative}.
            \item If a line has a \textit{random} category, then this is the only category for the line, and the main category is \textit{random}.
        \end{itemize}
    \end{itemize}

    \item \textbf{Was the “raw” data saved in addition to the preprocessed/cleaned/labeled data?}
    \begin{itemize}
        \item No.
    \end{itemize}

    \item \textbf{Is the software used to preprocess/clean/label the instances available?}
    \begin{itemize}
        \item \qXXXVII
    \end{itemize}

    \item \textbf{Any other comments?} 
    \begin{itemize}
        \item We provide a distribution of lines for each set and each category in Table~\ref{tab:code-completion-line-counts}.
        \begin{table*}[t]
            \centering
            \caption{Line counts for different sets in the project-level code completion dataset.}
            \vspace{0.2cm}
            \resizebox{\textwidth}{!}{
            \begin{tabular}{l  c c c c c c c c}
                \toprule
                  \multicolumn{1}{c}{\textbf{Set}} & \textit{\textbf{infile}} & \textit{\textbf{inproject}} & \textit{\textbf{common}} & \textit{\textbf{committed}} & \textit{\textbf{non-informative}} & \textit{\textbf{random}} & \textit{\textbf{all}} & \textbf{Avg. for one file}\\
                \midrule
                Small & 1,430 & 95 & 500 & 1,426 & 532 & 703 & 4,686 & 32.5 \\
                Medium & 2,224 & 2,236 & 779 & 1,495 & 858 & 1,084 & 8,676 & 38.7 \\
                Large & 2,691 & 2,595 & 693 & 1,322 & 1,019 & 1,311 & 9,631 & 35.7 \\
                Huge & 2,608 & 2,901 & 692 & 1,019 & 1,164 & 1,426 & 9,810 & 33.1 \\
                \bottomrule
            \end{tabular}
            }
            \label{tab:code-completion-line-counts}
        \end{table*}
    \end{itemize}
\end{enumerate}

\subsection{Uses}

\begin{enumerate}[start=39,label={Q\arabic*}]

    \item \textbf{Has the dataset been used for any tasks already?}
    \begin{itemize}
        \item We use the dataset to evaluate how well pre-trained code LLMs can utilize context from the given repository. We provide the evaluation results for CodeLlama 7B in Table~\ref{tab:completion_results} (see the \href{https://huggingface.co/spaces/JetBrains-Research/long-code-arena}{online leaderboard} for other models).
        \item We evaluate publicly available models as baselines without any modifications or fine-tuning. We implement several approaches to compose the context that fits into the model's context window (see \ref{code-completion-composers}). One of the best performing composers is the Path distance composer, for which the results are present in \Cref{tab:completion_results}. This composer chooses \texttt{.py} files from the repository snapshot that are in the same directory as the completion file or in the nearby directories, first picking files closer in the file tree to the completion file. As the context window sizes for all models are limited, we truncate the input sequence to the respective context size.
        \item We also report the results for the file-level context, which feeds the models only the prefix of the completion file for each completion line.

        \begin{table*}[t]
            \centering
            \caption{Results of the project-level code completion for CodeLlama 7B. The metric is Exact Match for the generated line.}
            \vspace{0.2cm}
            \resizebox{\textwidth}{!}{
            \begin{tabular}{l l c c c c c c c}
                \toprule
                 \multicolumn{1}{c}{\textbf{Set}} & \textbf{Context} & \textit{\textbf{infile}} & \textit{\textbf{inproject}} & \textit{\textbf{committed}} & \textit{\textbf{common}} & \textit{\textbf{non-informative}} & \textit{\textbf{random}} & \textit{\textbf{all}}\\
                 \midrule
                 \multirow{3}{*}{Small}
                 & File-level & 0.35 & 0.16 & 0.33 & 0.32 & 0.28 & 0.42 & 0.35 \\
                 & Path Distance 16K & 0.37 & 0.27 & 0.34 & 0.33 & 0.29 & 0.43 & 0.37 \\
                 \rowcolor{gray!30} & Difference & $+6\%$ & $\mathbf{+68\%}$ & $+3\%$ & $+3\%$ & $+2\%$ & $+2\%$ & $+5\%$ \\
                 \midrule
                 \multirow{3}{*}{Medium}
                 & File-level & 0.37 & 0.32 & 0.38 & 0.31 & 0.31 & 0.50 & 0.39\\
                 & Path Distance 16K & 0.43 & 0.49 & 0.42 & 0.44 & 0.44 & 0.58 & 0.49\\
                 \rowcolor{gray!30} & Difference & $+16\%$ & $\mathbf{+53\%}$ & $+10\%$ & $+42\%$ & $+42\%$ & $+16\%$ & $+26\%$ \\
                 \midrule
                 \multirow{3}{*}{Large}
                 & File-level & 0.36 & 0.29 & 0.39 & 0.34 & 0.30 & 0.44 & 0.35\\
                 & Path Distance 16K & 0.46 & 0.44 & 0.55 & 0.46 & 0.42 & 0.54 & 0.47\\
                 \rowcolor{gray!30} & Difference & $+27\%$ & $\mathbf{+52\%}$ & $+41\%$ & $+35\%$ & $+40\%$ & $+23\%$ & $+35\%$ \\
                 \midrule
                 \multirow{3}{*}{Huge}
                 & File-level & 0.40 & 0.34 & 0.44 & 0.34 & 0.30 & 0.50 & 0.39\\
                 & Path Distance 16K & 0.44 & 0.43 & 0.54 & 0.41 & 0.40 & 0.54 & 0.45\\
                 \rowcolor{gray!30} & Difference & $+10\%$ & $+26\%$ & $+22\%$ & $+20\%$ & $\mathbf{+36\%}$ & $+8\%$ & $+17\%$ \\
                 \bottomrule
            \end{tabular}
            }
            
            \label{tab:completion_results}
        \end{table*}
        
    \end{itemize}

    \item \textbf{Is there a repository that links to any or all papers or systems that use the dataset?}
    \begin{itemize}
        \item \qXL
    \end{itemize}

    \item \textbf{What tasks could the dataset be used for?}
    \begin{itemize}
        \item The provided dataset can be used in different tasks:
        \begin{itemize}
            \item to evaluate various approaches to utilize long context for code models, \eg retrieval-augmented generation, support of long context windows, etc.;
            \item to explore how code files in other languages or non-code files affect code completion;
            \item to compare benefits from long contexts with the associated increase in costs.
        \end{itemize} 
    \end{itemize}

    \item \textbf{Is there anything about the composition of the dataset or the way it was collected and preprocessed/cleaned/labeled that might impact future uses?}
    \begin{itemize}
        \item \qXLII
    \end{itemize}
    
    \item \textbf{Are there tasks for which the dataset should not be used?}
    \begin{itemize}
        \item We ask users of the datasets not to use the provided data for training.
    \end{itemize}

    \item \textbf{Any other comments?}\label{code-completion-composers}
    \begin{itemize}
        \item We provide several context composers as baselines.
        \begin{itemize}
            \item \textit{Naive composer} --- all the files from the repository snapshot are concatenated into one string with no specific order.
            \item \textit{Path distance composer} --- the order of the files is defined by the distance between files in a project file tree: if the file from the repository is closer to the completion file, then its content is closer in the context.
            \item \textit{File length composer} --- the order of the files is defined by the length of a file: shorter files are closer to the completion file.
            \item \textit{Half memory composer} --- each line from the repository files is removed with a probability of $0.5$, and the order of the files is the same as in the naive composer.
            \item \textit{Imports first composer} --- the order of the files is defined by an import relation of first degree: if any project files are imported in the completion file, then these files are closer to the completion file.
            \item \textit{Only declarations composer} --- some project files are left only with declaration lines, so we keep only names from the repository files.
        \end{itemize}
        
       \item  We leave further exploration of different context composers for future work. We present results for different context composers for CodeLlama 7B and the \textit{medium-context} dataset in Table~\ref{tab:code-completion-composers-choice}. Our experiments show that the perplexity values are different, but the order of composers performance is the same. A number in the column name means the maximum number of tokens in the context from the repository snapshot. 
    \end{itemize}
\end{enumerate}

\subsection{Distribution}

\begin{enumerate}[start=45,label={Q\arabic*}]
    \item \textbf{Will the dataset be distributed to third parties outside of the entity?}
    \begin{itemize}
        \item \qXLV
    \end{itemize}

    \item \textbf{How will the dataset be distributed? Does the dataset have a digital object identifier (DOI)?}
    \begin{itemize}
        \item The dataset is available through DOI at the HuggingFace Hub: {\url{https://doi.org/10.57967/hf/2512}}.
    \end{itemize}

\begin{table}[t]
    \centering
    \caption{The results for different context composers. The metric is perplexity on the completion file.}
    \resizebox{\textwidth}{!}{
    \begin{tabular}{l c c c c c c c}
        
        \toprule
        \multirow{2}{*}{\textbf{Additional context}}  & \multicolumn{3}{c}{\textbf{All files}} & \multicolumn{3}{c}{\textbf{Only Python files}} & \multirow{2}{*}{\textbf{Difference with FL}} \\\cmidrule(lr){2-4}\cmidrule(lr){5-7}
          & \textbf{256} & \textbf{1,753} & \textbf{12,000} & \textbf{256} & \textbf{1,753} & \textbf{12,000} & \\
        \midrule
       File-level (FL) & 1.849 & 1.849 & 1.849 & 1.849 & 1.849 & 1.849 & 0.000 \\ 
        Naive & 1.798 & 1.788 & 1.761 & 1.788 & 1.760 & 1.677 &  0.172\\
        Path distance (PD) & 1.783 & 1.727 & \textbf{1.607} & 1.782 & 1.726 & \textbf{1.601} & 0.248 \\
        Half memory (HM) & 1.799 & 1.789 & 1.743  & 1.789 & 1.765 & 1.670 & 0.179\\
       HM + PD & 1.782 & 1.730 & \underline{1.636} & 1.783 & 1.729 & \underline{1.636} & 0.213 \\
        File length & 1.797 & 1.784 & 1.742 & 1.792 & 1.774 & 1.708 & 0.141\\
        Imports First & 1.791 & 1.769 & 1.732 & 1.785 & 1.751 & 1.666 & 0.183\\
        Only declaration + PD\tablefootnote{We leave only declarations in all files except for one.} & 1.785 & 1.741 & 1.710 & 1.785 & 1.739 & 1.708 & 0.141\\
        \bottomrule
    \end{tabular}
    }
    \label{tab:code-completion-composers-choice}

\end{table}

    \item \textbf{When will the dataset be distributed?}
    \begin{itemize}
        \item \qXLVII
    \end{itemize}

    \item \textbf{Will the dataset be distributed under a copyright or other intellectual property (IP) license, and/or under applicable terms of use (ToU)?}
    \begin{itemize}
        \item \qXLVIII
    \end{itemize}

    \item \textbf{Have any third parties imposed IP-based or other restrictions on the data associated with the instances?}
    \begin{itemize}
        \item No.
    \end{itemize}

    \item \textbf{Do any export controls or other regulatory restrictions apply to the dataset or to individual instances?}
    \begin{itemize}
        \item \qL
    \end{itemize}

    \item \textbf{Any other comments?}
    \begin{itemize}
        \item No.
    \end{itemize}
\end{enumerate}

\subsection{Maintenance}
\begin{enumerate}[start=52,label={Q\arabic*}]
    \item \textbf{Who is supporting/hosting/maintaining the dataset?}
    \begin{itemize}
        \item \qLII
    \end{itemize}

    \item \textbf{How can the owner/curator/manager of the dataset be contacted (\textit{e.g.}, email address)?}
    \begin{itemize}
        \item \qLIII
    \end{itemize}

    \item \textbf{Is there an erratum?}
    \begin{itemize}
        \item \qLIV
    \end{itemize}

    \item \textbf{Will the dataset be updated? (\textit{e.g.}, to correct labeling errors, add new instances, delete instances)?}
    \begin{itemize}
        \item \qLV
    \end{itemize}

    \item \textbf{If the dataset relates to people, are there applicable limits on the retention of the data associated with the instances?}
    \begin{itemize}
        \item \qLVI
    \end{itemize}

    \item \textbf{Will older versions of the dataset continue to be supported/hosted/maintained?}
    \begin{itemize}
        \item \qLVII
    \end{itemize}

    \item \textbf{If others want to extend/augment/build on/contribute to the dataset, is there a mechanism for them to do so?}
    \begin{itemize}
        \item \qLVIII
    \end{itemize}

    \item \textbf{Any other comments?} 
    \begin{itemize}
        \item No.
    \end{itemize}
\end{enumerate}

\section{Datasheet for the Commit Message Generation dataset}
\label{sec:sheet4}

\subsection{Motivation}


\begin{enumerate}[start=1,label={Q\arabic*}]
    \item \textbf{For what purpose was the dataset created?}
    \begin{itemize}
        \item Commit message generation benchmark from Long Code Arena aims to evaluate machine learning models that generate natural language descriptions for large changes in software projects. Prior works on commit message generation typically address smaller changes and do not clean the data to the rigorous standards of manual curation.
    \end{itemize}
    
    \item \textbf{Who created this dataset (\textit{e.g.}, which team, research group) and on behalf of which entity (\textit{e.g.,} company, institution, organization)?}
    \begin{itemize}
        \item \qII
    \end{itemize}

    \item \textbf{Who funded the creation of the dataset?}
    \begin{itemize}
        \item \qIII
    \end{itemize}

    \item \textbf{Any other comments?}
    \begin{itemize}
        \item No.
    \end{itemize}
\end{enumerate}

\subsection{Composition}

\begin{enumerate}[start=5,label={Q\arabic*}]

    \item \textbf{What do the instances that comprise the dataset represent?}

    \begin{itemize}
        \item Each instance in the dataset represents a commit from a GitHub repository, with metadata like commit SHA and full repository name, ground truth commit message, and the list of performed changes in the Git diff format. Also, the dataset includes snapshots of all associated repositories to facilitate context construction. 
    \end{itemize}

    \item \textbf{How many instances are there in total (of each type, if appropriate)?}
    \begin{itemize}
        \item There are 163 datapoints in total.
    \end{itemize}

    \item \textbf{Does the dataset contain all possible instances or is it a sample (not necessarily random) of instances from a larger set?}
    \begin{itemize}
        \item The dataset is a sample from the test set of the CommitChronicle dataset, which is a vast collection of commits from GitHub repositories.
    \end{itemize}

    \item \textbf{What data does each instance consist of?}
    \begin{itemize}
        \item The structure of the datapoints is presented in \Cref{tab:cmg_dp_struct}.

\begin{table}[t]
\centering
\caption{The structure of datapoints in the commit message generation dataset.}
\label{tab:cmg_dp_struct}
\begin{tabular}{cp{6cm}}
\toprule
\textbf{Field} & \multicolumn{1}{c}{\textbf{Description}} \\
\midrule
\rowcolor{gray!30}\textbf{repo} & The full name of the GitHub repository the commit comes from \\
\textbf{hash} & The SHA hash of the commit, serves as an identifier inside individual repository \\
\rowcolor{gray!30}\textbf{date} & The timestamp of the commit (from the commit author) \\
\textbf{license} & The type of the license in the repository of the commit \\
\rowcolor{gray!30}\textbf{message} & The ground truth commit message \\
\textbf{mods} & The changes performed in a commit, represented as a list of per-file modifications, where the structure of a per-file modification is described in \Cref{tab:cmg_mods_struct} \\
\bottomrule
\end{tabular}
\end{table}

\begin{table}[t]
\centering
\caption{The structure of a per-file modification in the commit message generation dataset.}
\label{tab:cmg_mods_struct}
\begin{tabular}{cp{6cm}}
\toprule
\textbf{Field} & \multicolumn{1}{c}{\textbf{Description}} \\
\midrule
\rowcolor{gray!30}\textbf{change\_type} & The type of change to the current file, one of: ADD, COPY, RENAME, DELETE, MODIFY, or UNKNOWN\\
\textbf{old\_path} & The path to file before the change (might be empty if the file was created) \\
\rowcolor{gray!30}\textbf{new\_path} & The path to file after change (might be empty if the file was deleted) \\
\textbf{diff} & The changes to the current file, represented in a Git diff format \\
\bottomrule
\end{tabular}
\end{table}

    \end{itemize}

    \item \textbf{Is there a label or target associated with each instance?}
    \begin{itemize}
        \item The ground truth commit message for each commit can be regarded as target description for the changes from the corresponding commit.
    \end{itemize}

    \item \textbf{Is any information missing from individual instances?}
    \begin{itemize}
        \item No.
    \end{itemize}

    \item \textbf{Are relationships between individual instances made explicit?}
    \begin{itemize}
        \item \qXI
    \end{itemize}

    \item \textbf{Are there recommended data splits (\textit{e.g.}, training, development/validation, testing)?}
    \begin{itemize}
        \item \qXII
    \end{itemize}

    \item \textbf{Are there any errors, sources of noise, or redundancies in the dataset?}
    \begin{itemize}
        \item See preprocessing in \ref{q_d_35}.
    \end{itemize}

    \item \textbf{Is the dataset self-contained, or does it link to or otherwise rely on external resources?}
    \begin{itemize}
        \item The dataset is self-contained, as it provides the snapshots of all associated repositories.
    \end{itemize}

    \item \textbf{
    Does the dataset contain data that might be considered confidential?
    }
    \begin{itemize}
        \item \qXV
    \end{itemize}

    \item \textbf{Does the dataset contain data that, if viewed directly, might be offensive, insulting, threatening, or might otherwise cause anxiety?}
    \begin{itemize}
        \item \qXVI
    \end{itemize}

    \item \textbf{Does the dataset relate to people?}

    \begin{itemize}
        \item \qXVII
    \end{itemize}

    \item \textbf{Does the dataset identify any subpopulations (\textit{e.g.}, by age, gender)?}
    \begin{itemize}
        \item \qXVIII
    \end{itemize}

    \item \textbf{Is it possible to identify individuals (\textit{i.e.}, one or more natural persons), either directly or indirectly (\textit{i.e.}, in combination with other data) from the dataset?}
    \begin{itemize}
        \item \qXIX
    \end{itemize}

    \item \textbf{Does the dataset contain data that might be considered sensitive in any way?}
    \begin{itemize}
        \item \qXX
    \end{itemize}

    \item \textbf{Any other comments?} 
    \begin{itemize}
        \item No.
    \end{itemize}
\end{enumerate}

\subsection{Collection}

\begin{enumerate}[start=22,label={Q\arabic*}]
    \item \textbf{How was the data associated with each instance acquired?}
    \begin{itemize}
        \item The data associated with each instance was acquired directly from the CommitChronicle dataset~\cite{our-cmg-paper}.
    \end{itemize}

    \item \textbf{What mechanisms or procedures were used to collect the data (\textit{e.g.}, hardware apparatus or sensor, manual human curation, software program, software API)?}
    \begin{itemize}
        \item We refer the reader to the work of~\citet{our-cmg-paper} for the details about data collection. We also perform manual validation to select high-quality examples with long diffs and commit messages.
    \end{itemize}
    
    \item \textbf{
    If the dataset is a sample from a larger set, what was the sampling strategy?}
    \begin{itemize}
        \item This dataset is based on the test set of CommitChronicle to leave the disjoint train and validation sets for further experiments. We leave only Python commits from the test set and perform rigorous filtering as described in~\ref{q_d_35}.
    \end{itemize}

    \item \textbf{Who was involved in the data collection process (\textit{e.g.}, students, crowdworkers, contractors) and how were they compensated?}
    \begin{itemize}
        \item The data collection process was conducted by the authors of CommitChronicle, \citet{our-cmg-paper}.
    \end{itemize}

    \item \textbf{Over what timeframe was the data collected?}
    \begin{itemize}
        \item The CommitChronicle dataset~\cite{our-cmg-paper} was collected in February 2023. The construction of this dataset took place between October 2023 and January 2024.
    \end{itemize}

    \item \textbf{Were any ethical review processes conducted?}
    \begin{itemize}
        \item No.
    \end{itemize}

    \item \textbf{Does the dataset relate to people?}
    \begin{itemize}
        \item \qXXVIII
    \end{itemize}

    \item \textbf{Did you collect the data from the individuals in question directly, or obtain it via third parties or other sources (\textit{e.g.}, websites)?}
    \begin{itemize}
        \item \qXXIX
    \end{itemize}

    \item \textbf{Were the individuals in question notified about the data collection?}
    \begin{itemize}
        \item \qXXX
    \end{itemize}

    \item \textbf{Did the individuals in question consent to the collection and use of their data?}
    \begin{itemize}
        \item \qXXXI
    \end{itemize}

    \item \textbf{If consent was obtained, were the consenting individuals provided with a mechanism to revoke their consent in the future or for certain uses?}
    \begin{itemize}
        \item \qXXXII
    \end{itemize}

    \item \textbf{Has an analysis of the potential impact of the dataset and its use on data subjects been conducted?}
    \begin{itemize}
        \item \qXXXIII
    \end{itemize}

    \item \textbf{Any other comments?} 
    \begin{itemize}
        \item No.
    \end{itemize}
\end{enumerate}

\subsection{Preprocessing / Cleaning / Labeling}\label{sec:supp-cmg-preprocessing}

\begin{enumerate}[start=35,label={Q\arabic*}]
    \item \textbf{Was any preprocessing/cleaning/labeling of the data done?}
    \label{q_d_35}
    \begin{itemize}
        \item 
The exact data processing steps are listed in \Cref{tab:supp_cmg_dataset_filters}. For the commit message quality filter, we refine the dataset released in a recent study from Li and Ahmed~\cite{commit-msg-quality} to make it more suitable for data filtering purposes, and fine-tune the CodeBERT~\cite{codebert} model. Our commit message quality dataset\footnote{Commit message quality dataset: \url{https://huggingface.co/datasets/saridormi/commit-message-quality}} and classifier\footnote{Commit message quality classifier: \url{https://huggingface.co/saridormi/commit-message-quality-codebert}} are available online.

\begin{table}[t]
    \centering
    \caption{Filters applied to the CommitChronicle subset to build the commit message generation dataset from Long Code Arena. \textbf{*}Since the \textit{Quality} filter is based on a deep learning classifier, it was applied only to the subset of 3,366 commits obtained by running all the other filters.}
    \resizebox{\textwidth}{!}{
    \begin{tabular}{c p{3cm} p{7cm} p{3cm}}
        \toprule
         & \multicolumn{1}{c}{\textbf{Filter Description}} & \multicolumn{1}{c}{\textbf{Filter Details}} & \textbf{Number of commits rejected by the filter (\% of initial sample)}\\
         \midrule
         \multirow{4}{*}{\textbf{Diff Filters}}
         & \cellcolor{gray!30}Hash Diffs & \cellcolor{gray!30}Diff has whitespace-separated character-to-words ratio $\leq 20$~\cite{starcoder}. & \cellcolor{gray!30}437 (0.25\%) \\
         & Modification & Diff consists only of modifications of existing files (no additions,  deletions, renaming, or copying). & 25,750 (14.95\%) \\
         \midrule
         \multirow{15}{*}{\textbf{Message Filters}}
         & \cellcolor{gray!30}Capitalization & \cellcolor{gray!30}Message starts with an uppercase letter~\cite{octopack}. & \cellcolor{gray!30}68,384 (39.70\%) \\
         & Verbs & Message starts with any of the curated set of verbs from the recent work of~\citet{octopack}. & 90,696 (52.66\%) \\
         & \cellcolor{gray!30}References & \cellcolor{gray!30}Message does not contain external references (URLs or references to issues/pull requests). & \cellcolor{gray!30}31,487 (18.28\%) \\
         & Noise & Message does not follow certain patterns considered automatically generated or trivial~\cite{our-cmg-paper, octopack}. & 6,304 (3.66\%) \\
         & \cellcolor{gray!30}Min Words & \cellcolor{gray!30}Message contains $\geq 4$ words (whitespace-separated). & \cellcolor{gray!30}24,474 (14.21\%)\\
         & Min Lines & Message contains $\geq 2$ lines. & 138,160 (80.22\%)\\
         & \cellcolor{gray!30}Hash Messages & \cellcolor{gray!30}Message has whitespace-separated character-to-words ratio $\leq 20$~\cite{starcoder} and does not contain any SHA hashes~\cite{our-cmg-paper}. & \cellcolor{gray!30}12,540 (7.28\%) \\
         & Quality & Message is considered good by the commit message quality classifier. & 106 (3.14\%)*\\
         \bottomrule
    \end{tabular}
    }
    \label{tab:supp_cmg_dataset_filters}
\end{table}

\item After filtering, we retain 3,260 commits. Since we aim to target commits with larger changes, after the initial filtering, we only keep samples where the number of characters in diffs related to \texttt{.py} files is $\geq$ 3,000 characters. That leaves us with 858 commits that we further filter manually.

\item The manual labeling is conducted by one of the authors. We employ a 5-point Likert scale and additionally provide comments that elaborate on the reasoning for most of the samples. To facilitate further research, we made all the labels and comments available in the dataset.

    \end{itemize}

    \item \textbf{Was the “raw” data saved in addition to the preprocessed/cleaned/labeled data?}
    \begin{itemize}
        \item The raw data about each commit from the repositories included in the final version of our dataset can be obtained from the provided repositories' snapshots.
    \end{itemize}

    \item \textbf{Is the software used to preprocess/clean/label the instances available?}
    \begin{itemize}
        \item \qXXXVII
    \end{itemize}

    \item \textbf{Any other comments?} 
    \begin{itemize}
        \item No.
    \end{itemize}
\end{enumerate}

\subsection{Uses}\label{sec:supp-cmg-results}

\begin{enumerate}[start=39,label={Q\arabic*}]

    \item \textbf{Has the dataset been used for any tasks already?}
    \begin{itemize}
        \item We run multiple instruction-tuned LLMs on the presented commit message generation benchmark in a zero-shot setting (\emph{i.e.}, no examples in the prompt, only a natural language instruction). We employ the same prompt for all models, which we refine to address the most frequent issues in the generated messages from pilot experiments. The prompt is presented in Figure~\ref{fig:supp-cmg-prompt}. We only incorporate commit changes represented as diffs returned by the \texttt{git diff} command to prompt the LLMs and leave collection of more sophisticated contexts for future works. Additionally, we run the CodeT5~\cite{codet5} model fine-tuned for commit message generation task on the training part of the CommitChronicle dataset. This model only takes the commit diff as an input.

\begin{figure}[h]
\centering
\begin{BVerbatim}
Write a commit message for a given diff. Start with a heading that
serves as a summary of the whole diff: a single sentence in an
imperative form, no more than 50 characters long. If you have details
to add, do it after a blank line. Do your best to be specific, do not
use `refactor' unless you are absolutely sure that this change is ONLY
a refactoring. Your goal is to communicate what the change does
without having to look at the source code. Do not go into low-level
details like all the changed files, do not be overly verbose. Avoid
adding any external references like issue tags, URLs or emails. Diff:

[DIFF]

Commit message:
\end{BVerbatim}
\caption{The prompt for the commit message generation task.}\label{fig:supp-cmg-prompt}
\end{figure}

    \item We run each model three times with different random seeds and report average metrics across runs.
    We access OpenAI models through the official API. For all the other baselines, we use a single NVIDIA A100 GPU with default precision (except for Mixtral, where we use 8-bit precision~\cite{bitsandbytes-8bit}) 
    and FlashAttention-2~\cite{flash-attention-2} enabled.
    For all the models, we set the temperature to 0.8 and allow them to generate up to 512 tokens. This upper bound is mostly set due to practical considerations, as the maximum length of a commit message in our Commit Message Generation dataset is only 58 whitespace-separated words. 
    The results are presented in Table~\ref{tab:cmg_results}.

\begin{table}[t]
\centering
\caption{Results for the CMG benchmark from Long Code Arena. \textit{R} stands for \textit{ROUGE} metric, \textit{BS} stands for \textit{BERTScore} metric, where \textit{BS (norm.)} is the normalized version. All model categories are sorted by the \textit{ROUGE-1} metric. The best result in the category is highlighted in \textbf{bold}, and the second best result is \underline{underlined}. *CodeT5 is the only model fine-tuned for the CMG task as opposed to the zero-shot setting for the rest of the models.}
\resizebox{\textwidth}{!}{
\begin{tabular}{clrrrrrrr}
\toprule 
& \multicolumn{1}{c}{\textbf{Model}} & \multicolumn{1}{c}{\textbf{BLEU}} & \multicolumn{1}{c}{\textbf{ChrF}} & \multicolumn{1}{c}{\textbf{R-1}} & \multicolumn{1}{c}{\textbf{R-2}} & \multicolumn{1}{c}{\textbf{R-L}} & \multicolumn{1}{c}{\textbf{BS}} & \makecell[c]{\textbf{BS}\\\textbf{(norm.)}}\\
\midrule
\multirow{4}{*}{\textbf{Proprietary}} &
GPT-4 Turbo (1106) & \textbf{2.803} & \textbf{34.391} & \textbf{26.622} & \textbf{5.296} & \textbf{17.717} & \textbf{0.856} & \textbf{0.146} \\
 & GPT-4 (0613) & \underline{2.127} & \underline{32.624} & \underline{23.497} & \underline{5.217} & \underline{16.033} & 0.852 & 0.124\\
 & GPT-3.5 Turbo (0613) & 2.101 & 26.664 & 19.976 & 4.227 & 14.447 & 0.846 & 0.087\\
 & GPT-3.5 Turbo (1106) & 1.885 & 20.698 & 18.424 & 3.815 & 14.087 & \underline{0.854} & \underline{0.136}\\
\midrule
\multirow{3}{*}{\textbf{OSS (medium)}} &
Mixtral 8 bit (8x7B) & \textbf{2.189} & \textbf{31.984} & \textbf{23.61} & \textbf{5.376} & \textbf{16.329} & \textbf{0.848} & \textbf{0.097}\\
 & DeepSeek Coder (33B) & \underline{1.742} & \underline{29.08} & \underline{21.011} & \underline{4.471} & \underline{14.458} & 0.843 & 0.067\\
  & CodeLLaMA (34B) & 1.586 & 24.632 & 17.817 & 3.684 & 13.114 & \underline{0.844} & \underline{0.073}\\
\midrule
\multirow{4}{*}{\textbf{OSS (small)}} &
 Mistral (7B) & \textbf{1.895} & \textbf{30.719} & \textbf{23.648} & \textbf{4.458} & \textbf{16.262} & \textbf{0.847} & \textbf{0.096}\\
 & DeepSeek Coder (6.7B) & 1.634 & \underline{28.567} & \underline{20.188} & 3.604 & \underline{14.116} & 0.843 & 0.068\\
 & CodeLLaMA (13B) & \underline{1.727} & 23.099 & 18.207 & \underline{3.642} & 13.479 & \underline{0.844} & \underline{0.075}\\
& CodeLLaMA (7B) & 1.108 & 26.638 & 16.961 & 2.807 & 12.028 & 0.835 & 0.021\\
\midrule
\multirow{2}{*}{\textbf{OSS (tiny)}} & 
DeepSeek Coder (1.3B) & \textbf{0.75} & \textbf{22.449} & \textbf{13.815} & 2.029 & 9.753 & 0.822 & -0.057\\
& CodeT5* (220M) & 0.355 & 11.862 & 13.615 & \textbf{2.633} & \textbf{11.439} & \textbf{0.845} & \textbf{0.083}\\
\bottomrule
\end{tabular}
}
\label{tab:cmg_results}
\end{table}

    \end{itemize}

    \item \textbf{Is there a repository that links to any or all papers or systems that use the dataset?}
    \begin{itemize}
        \item \qXL
    \end{itemize}

    \item \textbf{What tasks could the dataset be used for?}
    \begin{itemize}
        \item The dataset can be directly employed for the commit message generation task. It might be used for other tasks related to the source code changes.
    \end{itemize}

    \item \textbf{Is there anything about the composition of the dataset or the way it was collected and preprocessed/cleaned/labeled that might impact future uses?}
    \begin{itemize}
        \item \qXLII
    \end{itemize}
    
    \item \textbf{Are there tasks for which the dataset should not be used?}
    \begin{itemize}
        \item \qXLIII
    \end{itemize}

    \item \textbf{Any other comments?}
    \begin{itemize}
        \item No.
    \end{itemize}
\end{enumerate}

\subsection{Distribution}

\begin{enumerate}[start=45,label={Q\arabic*}]
    \item \textbf{Will the dataset be distributed to third parties outside of the entity?}
    \begin{itemize}
        \item \qXLV
    \end{itemize}

    \item \textbf{How will the dataset be distributed? Does the dataset have a digital object identifier (DOI)?}
    \begin{itemize}
        \item The dataset is available through DOI at the HuggingFace Hub: {\url{https://doi.org/10.57967/hf/2513}}.
    \end{itemize}

    \item \textbf{When will the dataset be distributed?}
    \begin{itemize}
        \item \qXLVII
    \end{itemize}

    \item \textbf{Will the dataset be distributed under a copyright or other intellectual property (IP) license, and/or under applicable terms of use (ToU)?}
    \begin{itemize}
        \item \qXLVIII
    \end{itemize}

    \item \textbf{Have any third parties imposed IP-based or other restrictions on the data associated with the instances?}
    \begin{itemize}
        \item No.
    \end{itemize}

    \item \textbf{Do any export controls or other regulatory restrictions apply to the dataset or to individual instances?}
    \begin{itemize}
        \item \qL
    \end{itemize}

    \item \textbf{Any other comments?}
    \begin{itemize}
        \item No.
    \end{itemize}
\end{enumerate}

\subsection{Maintenance}
\begin{enumerate}[start=52,label={Q\arabic*}]
    \item \textbf{Who is supporting/hosting/maintaining the dataset?}
    \begin{itemize}
        \item \qLII
    \end{itemize}

    \item \textbf{How can the owner/curator/manager of the dataset be contacted (\textit{e.g.}, email address)?}
    \begin{itemize}
        \item \qLIII
    \end{itemize}

    \item \textbf{Is there an erratum?}
    \begin{itemize}
        \item \qLIV
    \end{itemize}

    \item \textbf{Will the dataset be updated? (\textit{e.g.}, to correct labeling errors, add new instances, delete instances)?}
    \begin{itemize}
        \item \qLV
    \end{itemize}

    \item \textbf{If the dataset relates to people, are there applicable limits on the retention of the data associated with the instances?}
    \begin{itemize}
        \item \qLVI
    \end{itemize}

    \item \textbf{Will older versions of the dataset continue to be supported/hosted/maintained?}
    \begin{itemize}
        \item \qLVII
    \end{itemize}

    \item \textbf{If others want to extend/augment/build on/contribute to the dataset, is there a mechanism for them to do so?}
    \begin{itemize}
        \item \qLVIII
    \end{itemize}

    \item \textbf{Any other comments?} 
    \begin{itemize}
        \item No.
    \end{itemize}
\end{enumerate}

\section{Datasheet for the Bug Localization dataset}
\label{sec:sheet5}

\subsection{Motivation}


\begin{enumerate}[start=1,label={Q\arabic*}]
    \item \textbf{For what purpose was the dataset created?}
    \begin{itemize}
        \item The bug localization benchmark is a part of the Long Code Arena that serves to evaluate models' abilities in locating files that should be changed given a bug description. The dataset includes real issues that describe bugs, together with the respective pull requests (PRs) that fix them. The model under evaluation takes a bug description and the repository state before the fix and then outputs the list of files that need to be changed.
    \end{itemize}
    
    \item \textbf{Who created this dataset (\textit{e.g.}, which team, research group) and on behalf of which entity (\textit{e.g.,} company, institution, organization)?}
    \begin{itemize}
        \item \qII
    \end{itemize}

    \item \textbf{Who funded the creation of the dataset?}
    \begin{itemize}
        \item \qIII
    \end{itemize}

    \item \textbf{Any other comments?}
    \begin{itemize}
        \item No.
    \end{itemize}
\end{enumerate}

\subsection{Composition}

\begin{enumerate}[start=5,label={Q\arabic*}]

    \item \textbf{What do the instances that comprise the dataset represent?}

    \begin{itemize}
        \item Each datapoint contains three key elements: the bug description, the state of the repository where the bug is reproducible, and the list of files that need to be modified to resolve the bug. The bug description represents the body of the issue that was assigned a bug-related label. The repository state is represented by the commit SHA. The list of files that should be modified comes from the pull request that resolves the respective bug report.
    \end{itemize}

    \item \textbf{How many instances are there in total (of each type, if appropriate)?}
    \begin{itemize}
        \item The dataset contains 7,479 datapoints in total divided, between three sets by language:
         \begin{itemize}
            \item \texttt{py} --- change contains only Python files (4,339 datapoints);
            \item \texttt{java} --- change contains only Java files (2,522 datapoints);
            \item \texttt{kt} --- change contains only Kotlin files (618 datapoints).
        \end{itemize}
        50 datapoints for each language are manually verified in order to form a test subset for model evaluation (150 datapoints in total).
    \end{itemize}
    \item \textbf{Does the dataset contain all possible instances or is it a sample (not necessarily random) of instances from a larger set?}
    \begin{itemize}
        \item \qVII It comes from a larger set of issues in Python, Kotlin, and Java GitHub repositories.
    \end{itemize}

    \item \textbf{What data does each instance consist of?}
    \begin{itemize}
        \item The core fields in the datapoints are presented in Table \ref{tab:bl_fields_description}. 
    
    \begin{table}[t]
    \centering
    \caption{Description of datapoints in the bug localization dataset.}
    \begin{tabular}{cp{9cm}}
    \toprule
    \textbf{Field} & \multicolumn{1}{c}{\textbf{Description}} \\
    \midrule
    \rowcolor{gray!30}\textbf{id} & Datapoint ID \\
     \textbf{repo\_owner} & Bug issue repository owner \\
     \rowcolor{gray!30}\textbf{repo\_name} & Bug issue repository name \\
    \textbf{static\_id} & Datapoint text ID \\
   \rowcolor{gray!30}  \textbf{issue\_url} & GitHub link to issue \\
    \textbf{issue\_title} & Issue title \\
    \rowcolor{gray!30}\textbf{issue\_body} & Issue body with bug description \\
     \textbf{issue\_labels} & List of labels assigned to issue \\
   \rowcolor{gray!30} \textbf{pull\_url} & GitHub link to PR \\
     \textbf{pull\_create\_at} & Date of PR creation in format of \texttt{yyyy-mm-ddThh:mm:ssZ} \\
    \rowcolor{gray!30}\textbf{base\_sha} & PR base SHA \\
     \textbf{head\_sha} & PR head SHA \\
    \rowcolor{gray!30} \textbf{diff\_url} & PR diff URL between base and head SHA \\
    \textbf{diff} & PR diff content \\
    \rowcolor{gray!30}\textbf{changed\_files} & List of changed files parsed from diff \\
     \textbf{link\_url} & GitHub link to issue or PR comment from which the link was parsed \\
    \rowcolor{gray!30}\textbf{links\_count} & Number of links between the issue and the PR,  equals \texttt{2} if the link is mutual, \texttt{1} if it is one-sided \\
     \textbf{link\_keyword} & "Fix"-related keyword which surrounds the issue link \\
     \rowcolor{gray!30}\textbf{stars} & Number of repository stars \\
    \textbf{language} & Main programming language for repository \\
    \bottomrule
    \end{tabular}
    \label{tab:bl_fields_description}
    \end{table}

    \item Based on the core fields, we calculated the number of statistics and attached them to each datapoint. The additional fields are presented in Table~\ref{tab:bl_metrics_fields_description}. We excluded test files from the experiment because their modifications typically only support program repairs and do not contain the actual bugs. Thus, all metrics are calculated on all project files except for the test files.
    
    \begin{table}[h]
    \centering
    \caption{Description of additional metrics calculated on the bug localization dataset.}
    \begin{tabular}{cp{6cm}}
    \toprule
    \textbf{Metric} & \multicolumn{1}{c}{\textbf{Description}} \\
    \midrule
    \rowcolor{gray!30}\textbf{issue\_symbols\_count} & Number of symbols in issue description \\
    \textbf{issue\_tokens\_count} & Number of tokens in issue description \\
    \rowcolor{gray!30}\textbf{issue\_words\_count} & Number of words in issue description \\
    
    \textbf{issue\_lines\_count} & Number of lines in issue description \\
    \rowcolor{gray!30}\textbf{issue\_code\_blocks\_count} & Number of triple quotes blocks parsed in issue description \\
    
    \textbf{issue\_links\_count} & Number of links parsed in issue description \\
    \midrule
    \rowcolor{gray!30}\textbf{diff\_symbols\_count} & Number of symbols in diff \\
    
    \textbf{diff\_tokens\_count} & Number of tokens in diff \\
    \rowcolor{gray!30}\textbf{diff\_words\_count} & Number of words in diff \\
    
    \textbf{issue\_lines\_count} & Number of lines in diff \\
    \rowcolor{gray!30}\textbf{changed\_files\_count} & Number of all changed files mentioned in diff \\
    
    \textbf{changed\_files\_without\_test\_count} & Number of changed files not including test files mentioned in diff \\
    \rowcolor{gray!30}\textbf{code\_changed\_files\_count} & Number of files written in Python, Java, or Kotlin mentioned in diff \\
    
    \textbf{py\_changed\_files\_count} & Number of Python files mentioned as changed in diff \\
    \rowcolor{gray!30}\textbf{java\_changed\_files\_count} & Number of Java files mentioned as changed in diff \\
    
    \textbf{kt\_changed\_files\_count} & Number of Kotlin files mentioned as changed in diff \\
    \midrule
    \rowcolor{gray!30}\textbf{repo\_symbols\_count} & Total number of symbols in repository's files \\
    
    \textbf{repo\_tokens\_count} & Total number of tokens in repository's files. \\
    \rowcolor{gray!30}\textbf{repo\_words\_count} & Total number of words in repository's files \\
    
    \textbf{repo\_lines\_count} & Total number of lines in repository's files \\
    \rowcolor{gray!30}\textbf{repo\_files\_count} & Total number of files in repository \\
    
    \textbf{repo\_files\_without\_test\_count} & Total number of files without tests in the repository \\
    \bottomrule
    \end{tabular}
    \label{tab:bl_metrics_fields_description}
    \end{table}

    \end{itemize}

    \item \textbf{Is there a label or target associated with each instance?}
    \begin{itemize}
        \item The target for the bug localization task is the list of files that should be changed (field \textbf{changed\_files} in the dataset).
    \end{itemize}

    \item \textbf{Is any information missing from individual instances?}
    \begin{itemize}
        \item No.
    \end{itemize}

    \item \textbf{Are relationships between individual instances made explicit?}
    \begin{itemize}
        \item \qXI
    \end{itemize}

    \item \textbf{Are there recommended data splits (\textit{e.g.}, training, development/validation, testing)?}
    \begin{itemize}
        \item The dataset contains the dedicated test split consisting of 150 examples that are manually verified for correctness. Along with it, we present a development split that was not manually checked but can be used by researchers for model development.
    \end{itemize}

    \item \textbf{Are there any errors, sources of noise, or redundancies in the dataset?}
    \begin{itemize}
        \item We describe the data collection process in~\ref{q_e_22}.
    \end{itemize}

    \item \textbf{Is the dataset self-contained, or does it link to or otherwise rely on external resources?}
    \begin{itemize}
        \item The dataset does not store the repository snapshots but rather fetches them from GitHub during benchmarking to reduce the dataset’s memory requirements.
    \end{itemize}

    \item \textbf{
    Does the dataset contain data that might be considered confidential?
    }
    \begin{itemize}
        \item \qXV
    \end{itemize}

    \item \textbf{Does the dataset contain data that, if viewed directly, might be offensive, insulting, threatening, or might otherwise cause anxiety?}
    \begin{itemize}
        \item \qXVI
    \end{itemize}

    \item \textbf{Does the dataset relate to people?}

    \begin{itemize}
        \item \qXVII
    \end{itemize}

    \item \textbf{Does the dataset identify any subpopulations (\textit{e.g.}, by age, gender)?}
    \begin{itemize}
        \item \qXVIII
    \end{itemize}

    \item \textbf{Is it possible to identify individuals (\textit{i.e.}, one or more natural persons), either directly or indirectly (\textit{i.e.}, in combination with other data) from the dataset?}
    \begin{itemize}
        \item \qXIX
    \end{itemize}

    \item \textbf{Does the dataset contain data that might be considered sensitive in any way?}
    \begin{itemize}
        \item \qXX
    \end{itemize}

    \item \textbf{Any other comments?} 
    \begin{itemize}
        \item No.
    \end{itemize}
\end{enumerate}

\subsection{Collection}

\begin{enumerate}[start=22,label={Q\arabic*}]
    \item \textbf{How was the data associated with each instance acquired?}
    \label{q_e_22}
\begin{itemize}
    \item To collect the data, we use the following protocol:

\begin{enumerate}

\item We start with the common corpus of collected GitHub repositories. Then, for each repository, we download information about all issues, pull requests, and comments using the GitHub API. As a result, we download more than 8M issues, 7M pull requests, and 34.4M comments.

\item GitHub API does not provide information about relations between issues and pull requests. We obtain these relations by parsing references from descriptions or comments. To do so, we write regular expressions for extracting all possible referencing formats as provided in GitHub documentation. To also collect the context around the reference, we capture one ``fix''-related keyword (\textit{e.g.}, \texttt{close}, \texttt{closes}, \texttt{closed}, \texttt{fix}, \texttt{fixes}, \texttt{fixed}, \texttt{resolve}, \texttt{resolves}, \texttt{resolved}, \texttt{solve}, \texttt{solves}, \texttt{solved}) before and after the link with the regular expressions.
We also check if references are mutual (if the issue refers to the pull request and vice versa) or not (if only a single link from either the issue or the pull request exists).

\item We sort all issue-PR pairs by the number of stars in the respective repository and assign each pair an \texttt{ID} based on its index in the sorted order. 
We populate the \texttt{diff} field by running a git command in a locally cloned repository to get the diff in a text format. Unfortunately, this method does not work for pull requests created from forks, so we save a null value for such cases.
\end{enumerate}
\end{itemize}

    \item \textbf{What mechanisms or procedures were used to collect the data (\textit{e.g.}, hardware apparatus or sensor, manual human curation, software program, software API)?}
    \begin{itemize}
        \item The data collection step used GitHub API. Then, we performed manual verification and assessment to select and filter data.
    \end{itemize}
    
    \item \textbf{
    If the dataset is a sample from a larger set, what was the sampling strategy?}
    \begin{itemize}
        \item The dataset is sampled from a larger set of issues and pull requests as described in \ref{q_e_22}.
    \end{itemize}

    \item \textbf{Who was involved in the data collection process (\textit{e.g.}, students, crowdworkers, contractors) and how were they compensated?}
    \begin{itemize}
        \item \qXXV
    \end{itemize}

    \item \textbf{Over what timeframe was the data collected?}
    \begin{itemize}
        \item The construction of this dataset took place between October 2023 and January 2024.
    \end{itemize}

    \item \textbf{Were any ethical review processes conducted?}
    \begin{itemize}
        \item No.
    \end{itemize}

    \item \textbf{Does the dataset relate to people?}
    \begin{itemize}
        \item \qXXVIII
    \end{itemize}

    \item \textbf{Did you collect the data from the individuals in question directly, or obtain it via third parties or other sources (\textit{e.g.}, websites)?}
    \begin{itemize}
        \item \qXXIX
    \end{itemize}

    \item \textbf{Were the individuals in question notified about the data collection?}
    \begin{itemize}
        \item \qXXX
    \end{itemize}

    \item \textbf{Did the individuals in question consent to the collection and use of their data?}
    \begin{itemize}
        \item \qXXXI
    \end{itemize}

    \item \textbf{If consent was obtained, were the consenting individuals provided with a mechanism to revoke their consent in the future or for certain uses?}
    \begin{itemize}
        \item \qXXXII
    \end{itemize}

    \item \textbf{Has an analysis of the potential impact of the dataset and its use on data subjects been conducted?}
    \begin{itemize}
        \item \qXXXIII
    \end{itemize}

    \item \textbf{Any other comments?} 
    \begin{itemize}
        \item No.
    \end{itemize}
\end{enumerate}

\subsection{Preprocessing / Cleaning / Labeling}

\begin{enumerate}[start=35,label={Q\arabic*}]
    \item \textbf{Was any preprocessing/cleaning/labeling of the data done?}
    \begin{itemize}
        \item To enhance the quality of our data, first, we apply empirical filters based on the fields from the dataset, as listed in \Cref{tab:dataset_filters}. Firstly, we retain only issues with ``bug'' mentioned in the labels and non-empty descriptions. Additionally, we remove issues containing links to media, as they may include crucial data visualizations that are inaccessible through other means. To ensure that most models can use the dataset for evaluation, we only keep issues written in English.
        For pull requests, we filter out those introducing new files and retain only pull requests modifying existing files, provided their diffs could be extracted from the cloned repository. Furthermore, to facilitate the future manual labeling process, we leave only pull requests written in Python, Java, or Kotlin, as these are languages known well to authors. To work with diffs and patches, as well as to extract the changed files and their modification modes, we use the unidiff package.\footnote{Undiff: \url{https://pypi.org/project/unidiff/}} Additionally, we avoid pull requests that include changes to media files with non-UTF-8 encoding, as such changes are often difficult to reproduce. The most crucial filter ensures that each pull request is associated with exactly one issue, and vice versa, to maintain the relevance of changes to issue descriptions and to prevent situations where a pull request addresses multiple issues or an issue is fixed by several pull requests. Following these filtration steps, 10,195 datapoints remain in the dataset.

    \begin{table}[t]
    \centering
    \caption{Empirical filters applied to the bug localization dataset.}
    \label{tab:dataset_filters}
    \begin{tabular}{c p{7cm} p{3cm}}
    \toprule
    \multicolumn{1}{c}{\textbf{Field}} & \multicolumn{1}{c}{\textbf{Description}} & \textbf{Number of datapoints rejected by the filter (\% of the initial set)} \\
    \midrule
    \rowcolor{gray!30}
    \textbf{issue\_labels} & At least one label should include "bug" as a substring & 3,472,057 (79.8\%) \\
    \textbf{issue\_body} & Description should not be empty & 16,265 (0.37\%) \\
    \rowcolor{gray!30}
    \textbf{issue\_body} & Description should contain only text without attached media & 145,225 (3.34\%) \\
    \textbf{issue\_body} & Description should be written mostly in English & 35,942 (0.83\%) \\
    \midrule
    \rowcolor{gray!30}
    \textbf{diff} & Diff can be extracted and should not be empty or corrupted & 475,447 (10.93\%) \\
    \textbf{diff} & Diff should consist only of modifications of existing files and no introduction of new files & 30,572 (0.7\%) \\
    \rowcolor{gray!30}
    \textbf{diff} & Diff should include at least one file in either Python, Java, or Kotlin & 138,653 (3.19\%) \\
    \textbf{diff} & Diff should include only UTF-8 files to filter out unreadable or graphical objects & 18 ($\le$ 0.01\%) \\
    \rowcolor{gray!30}\
    \textbf{base\_commit} & Repository content on base commit can be extracted and should not be empty or corrupted & 6,198 (0.14\%) \\
    \midrule
    \textbf{pull\_url} & PR should refer to no more than one issue & 7,376 (0.17\%) \\
    \rowcolor{gray!30}
    \textbf{issue\_url} & Issue should refer to no more than one pull request & 1,934 (0.04\%) \\
    \textbf{link\_keyword} & "fix"-related keyword should stay before or after link in the issue description. & 10,406 (0.24\%) \\
    \bottomrule
    \end{tabular}
    \end{table}

     \item On top of the previous filtering step, we remove outliers for several numerical fields, including \texttt{changed\_files\_count}, \texttt{changed\_lines\_count}, and \texttt{issue\_tokens\_count}. \Cref{tab:outliers_filters} shows the result of removing outliers.

    \begin{table}[t]
    \centering
    \caption{Outlier filters applied to the bug localization dataset.}
    \label{tab:outliers_filters}
    \begin{tabular}{c p{7cm} p{3cm}}
    \toprule
    \textbf{Field} & \multicolumn{1}{c}{\textbf{Description}} & \textbf{Number of datapoints rejected by the filter (\% of initial set)} \\
    \midrule
    \rowcolor{gray!30}
    \textbf{changed\_files\_count} & Number of changed files should not be more than 22 (0.99 quantile) & 100 ($\le$ 0.01\%) \\
    \textbf{changed\_lines\_count} & Number of changed lines should not be more than 594 (0.99 quantile) & 102 ($\le$ 0.01\%) \\
    \rowcolor{gray!30}
    \textbf{issue\_tokens\_count} & Issue description can be tokenized using GPT-4 tokenizer & 43 ($\le$ 0.01\%) \\
    \textbf{issue\_tokens\_count} & Issue description should contain at least 13 tokens (0.01 quantile) & 85 ($\le$ 0.01\%) \\
    \rowcolor{gray!30}
    \textbf{issue\_tokens\_count} & Issue description should contain no more than 4,500 tokens (0.99 quantile) & 103 ($\le$ 0.01\%) \\
    \bottomrule
    \end{tabular}
    \end{table}

    \item After data filtration, we are left with 7,479 datapoints that comprise the entire dataset. \Cref{tab:stats} presents statistics of the dataset, with the difference in statistics between languages being negligible. 

    \begin{table}[t]
    \centering
    \caption{Final statistics of the dataset.}
    \label{tab:stats}
    \begin{tabular}{c r r r r}
    \toprule
    \textbf{Field} & \multicolumn{1}{c}{\textbf{Min}} & \multicolumn{1}{c}{\textbf{Median}} & \multicolumn{1}{c}{\textbf{Mean}} & \multicolumn{1}{c}{\textbf{Max}} \\
    \midrule
    \textbf{repo\_files\_count} & 16 & 331 & 1,077 & 33,644 \\
    \textbf{repo\_lines\_count} & 9 & 52,743 & 145,377 & 8,687,912 \\
    \textbf{repo\_tokens\_count} & 78 & 488,286 & 1,684,619 & 225,649,725 \\
    \midrule
    \textbf{changed\_files\_count} & 1 & 1 & 2 & 21 \\
    \textbf{changed\_lines\_count} & 1 & 15 & 37 & 594 \\
    \textbf{changed\_tokens\_count} & 1 & 158 & 608 & 837,626 \\
    \midrule
    \textbf{issue\_words\_count} & 1 & 106 & 149 & 1,806 \\
    \textbf{issue\_lines\_count} & 1 & 22 & 33 & 586 \\
    \textbf{issue\_tokens\_count} & 13 & 227 & 432 & 4,491 \\
    \midrule
    \textbf{issue\_links\_count} & 0 & 0 & 0.80 & 56 \\
    \textbf{issue\_code\_blocks\_count} & 0 & 1 & 0.99 & 31 \\
    \bottomrule
    \end{tabular}
    \end{table}

    \item After the analysis of the dataset, we carry out manual data labeling and verification process to select the subset of high-quality datapoints for evaluation. First, we sort the datapoints by the number of stars in the respective repositories, assuming that popular repositories have better processes and quality for issue tracking and bug reporting. Then, we go through datapoints of each repository, selecting ones that meet the following criteria:
    \begin{itemize}
    \item The issue describes a single bug completely and exhaustively.
    \item The pull request is linked to the issue and resolves this issue alone.
    \item All changes are relevant to the described issue, with no extra functionality or side refactorings included.
    \item The changes were reviewed and accepted.
    \end{itemize}
    \item If a datapoint does not meet these criteria, we go to another one from the same repository, or if none are left, we move on to the next repository by the number of stars, until we select 50 good datapoints per language. To keep the distribution of the number of changed files, for each repository, we try to pick one datapoint with a single changed file and one datapoint with two or more changed files. This strategy allows us to collect a diverse set of datapoints from different repositories and keep the distribution of the number of changed files similar to the complete set of issues.
    \end{itemize}
        \item \textbf{Was the “raw” data saved in addition to the preprocessed/cleaned/labeled data?}
    \begin{itemize}
        \item No.
    \end{itemize}

    \item \textbf{Is the software used to preprocess/clean/label the instances available?}
    \begin{itemize}
        \item \qXXXVII
    \end{itemize}

    \item \textbf{Any other comments?} 
    \begin{itemize}
        \item No.
    \end{itemize}
\end{enumerate}

\subsection{Uses}

\begin{enumerate}[start=39,label={Q\arabic*}]

    \item \textbf{Has the dataset been used for any tasks already?}
    \begin{itemize}
        \item We run several baseline solutions on the bug localization task that utilize the presented dataset. The results are presented in~\Cref{results_bug_fixing}.
        \item First, we evaluate several retrieval-based approaches. The logic is straightforward: data analysis indicates that issue descriptions often include code blocks and stack traces pointing to the code responsible for bugs. Consequently, these descriptions should closely match the content of the files that require modification. Following this logic, we can compute embeddings for the bug report and all project files, and then identify project files that require fixing as the closest to the bug description by the cosine distance in the embedding space. We try several approaches to compute embeddings: TF-IDF with a BPE tokenizer pre-trained on the repository code, CodeT5~\cite{codet5}, CodeBERT~\cite{codebert}, GTE~\cite{li2023general}, and Mistral~\cite{mistral} models. Also, we evaluate BM25~\cite{bm25}, a classic approach from the information retrieval field.
        \item Second, we evaluate GPT-3.5 and GPT-4, prompting them to identify one to five bugged files using the bug description and the list of repository files. Figure~\ref{fig:bug-localization-prompt} presents the full prompt with placeholders. If the prompt exceeds the context size, we divide the file list into several queries. The final query combines all outputs and requests the final list sorted by relevance.
        \item For metrics, we calculate Recall@1 for the datapoints with one changed file, Recall@2 and Precision@2 for datapoints with two or more changed files. For all datapoints, we also calculate the F1-score and MAP, which we consider the target metric for model comparison. 

        \begin{figure}[t]
        \centering
\begin{BVerbatim}  
List of files: [FILES_LIST]
Issue: [ISSUE_TITLE] [ISSUE_DESCRIPTION]
You are given a list of files in the project and a bug issue
description. Select a subset of 1--5 files that SHOULD be fixed
according to the issue. Provide output in JSON format with one field
`files' which contains a list of file names that SHOULD be fixed.
Provide ONLY JSON without any additional comments.
\end{BVerbatim}
        \caption{Prompt for bug localization by GPT models.}
        \label{fig:bug-localization-prompt}
        \end{figure}
        
        \begin{table}[t]
        \centering
        \caption{The baseline results for the bug localization task.}
        \label{tab}
        \begin{tabular}{l c c c c c}
        \toprule
        \multicolumn{1}{c}{\textbf{Model}} & \textbf{R@1} & \textbf{R@2} & \textbf{P@2} & \textbf{F1-score} & \textbf{MAP} \\
        \midrule
        \textbf{TF-IDF+NLTK} & 0.16 & 0.1 & 0.15 & 0.13 & 0.20 \\
        \textbf{TF-IDF+BPE} & 0.30 & 0.15 & 0.24 & 0.21 & 0.28 \\
        \textbf{BM25} & 0.17 & 0.12 & 0.19 & 0.19 & 0.21 \\
        \textbf{CodeT5} & 0.28 & 0.13 & 0.17 & 0.18 & 0.23 \\
        \textbf{CodeBERT} & 0.29 & 0.15 & 0.18 & 0.20 & 0.25 \\
        \textbf{GTE} & 0.37 & 0.17 & 0.26 & 0.25 & 0.33 \\
        \textbf{Mistral} & 0.35 & 0.17 & 0.24 & 0.25 & 0.30 \\
        \midrule
        \textbf{GPT-3.5} & 0.49 & 0.19 & 0.31 & 0.35 & 0.29 \\
        \textbf{GPT-4} & \textbf{0.74} & \textbf{0.20} & \textbf{0.32} & \textbf{0.44} & \textbf{0.39} \\\bottomrule
        \end{tabular}
        \label{results_bug_fixing}
        \end{table}
        
    \end{itemize}

    \item \textbf{Is there a repository that links to any or all papers or systems that use the dataset?}
    \begin{itemize}
        \item \qXL
    \end{itemize}

    \item \textbf{What tasks could the dataset be used for?}
    \begin{itemize}
        \item The provided dataset can be used for evaluating bug localization approaches and other tasks related to bug-fixing problems.

    \end{itemize}

    \item \textbf{Is there anything about the composition of the dataset or the way it was collected and preprocessed/cleaned/labeled that might impact future uses?}
    \begin{itemize}
        \item \qXLII
    \end{itemize}
    
    \item \textbf{Are there tasks for which the dataset should not be used?}
    \begin{itemize}
        \item \qXLIII
    \end{itemize}

    \item \textbf{Any other comments?}
    \begin{itemize}
        \item No.
    \end{itemize}
\end{enumerate}

\subsection{Distribution}

\begin{enumerate}[start=45,label={Q\arabic*}]
    \item \textbf{Will the dataset be distributed to third parties outside of the entity?}
    \begin{itemize}
        \item \qXLV
    \end{itemize}

    \item \textbf{How will the dataset be distributed? Does the dataset have a digital object identifier (DOI)?}
    \begin{itemize}
        \item The dataset is available through DOI at the HuggingFace Hub: {\url{https://doi.org/10.57967/hf/2514}}.
    \end{itemize}

    \item \textbf{When will the dataset be distributed?}
    \begin{itemize}
        \item \qXLVII
    \end{itemize}

    \item \textbf{Will the dataset be distributed under a copyright or other intellectual property (IP) license, and/or under applicable terms of use (ToU)?}
    \begin{itemize}
        \item \qXLVIII
    \end{itemize}

    \item \textbf{Have any third parties imposed IP-based or other restrictions on the data associated with the instances?}
    \begin{itemize}
        \item No.
    \end{itemize}

    \item \textbf{Do any export controls or other regulatory restrictions apply to the dataset or to individual instances?}
    \begin{itemize}
        \item \qL
    \end{itemize}

    \item \textbf{Any other comments?}
    \begin{itemize}
        \item No.
    \end{itemize}
\end{enumerate}

\subsection{Maintenance}
\begin{enumerate}[start=52,label={Q\arabic*}]
    \item \textbf{Who is supporting/hosting/maintaining the dataset?}
    \begin{itemize}
        \item \qLII 
    \end{itemize}

    \item \textbf{How can the owner/curator/manager of the dataset be contacted (\textit{e.g.}, email address)?}
    \begin{itemize}
        \item \qLIII
    \end{itemize}

    \item \textbf{Is there an erratum?}
    \begin{itemize}
        \item \qLIV
    \end{itemize}

    \item \textbf{Will the dataset be updated? (\textit{e.g.}, to correct labeling errors, add new instances, delete instances)?}
    \begin{itemize}
        \item \qLV
    \end{itemize}

    \item \textbf{If the dataset relates to people, are there applicable limits on the retention of the data associated with the instances?}
    \begin{itemize}
        \item \qLVI
    \end{itemize}

    \item \textbf{Will older versions of the dataset continue to be supported/hosted/maintained?}
    \begin{itemize}
        \item \qLVII
    \end{itemize}

    \item \textbf{If others want to extend/augment/build on/contribute to the dataset, is there a mechanism for them to do so?}
    \begin{itemize}
        \item \qLVIII 
    \end{itemize}

    \item \textbf{Any other comments?} 
    \begin{itemize}
        \item No.
    \end{itemize}
\end{enumerate}

\section{Datasheet for the Module Summarization dataset}
\label{sec:sheet6}

\subsection{Motivation}


\begin{enumerate}[start=1,label={Q\arabic*}]
    \item \textbf{For what purpose was the dataset created?}
    \begin{itemize}
        \item Module summarization dataset is a part of the Long Code Arena that aims at testing models in generating documentation files. The minimal set of data for the task consists of an intent behind the documentation and the relevant part of the codebase. Based on the provided data, the model has to generate the documentation. The testing then happens by running an LLM assessor to decide which documentation is better: the generated one or the ground truth.
    \end{itemize}
    
    \item \textbf{Who created this dataset (\textit{e.g.}, which team, research group) and on behalf of which entity (\textit{e.g.,} company, institution, organization)?}
    \begin{itemize}
        \item \qII
    \end{itemize}

    \item \textbf{Who funded the creation of the dataset?}
    \begin{itemize}
        \item \qIII
    \end{itemize}

    \item \textbf{Any other comments?}
    \begin{itemize}
        \item No.
    \end{itemize}
\end{enumerate}

\subsection{Composition}

\begin{enumerate}[start=5,label={Q\arabic*}]

    \item \textbf{What do the instances that comprise the dataset represent?}

    \begin{itemize}
        \item Each instance in the dataset represents an instruction for generating a documentation file (intent and name of the original file), as well as a snapshot of a repository that the model should use for generation. \Cref{tab:m2t_dp_struct} shows the detailed structure of the datapoints. 
    \end{itemize}

    \item \textbf{How many instances are there in total (of each type, if appropriate)?}
    \begin{itemize}
        \item There are 216 datapoints in total.
    \end{itemize}

    \item \textbf{Does the dataset contain all possible instances or is it a sample (not necessarily random) of instances from a larger set?}
    \begin{itemize}
        \item \qVII  It comes from a larger set of Python repositories.
    \end{itemize}

    \item \textbf{What data does each instance consist of?}
    \begin{itemize}
        \item The structure of the datapoints is presented in Table \ref{tab:m2t_dp_struct}.
    \end{itemize}

    \begin{table}[t]
    \centering
    \caption{The structure of datapoints in the module summarization dataset.}
    \label{tab:m2t_dp_struct}
    \begin{tabular}{cp{6cm}}
    \toprule
    \textbf{Field} & \multicolumn{1}{c}{\textbf{Description}} \\
    \midrule
    \rowcolor{gray!30}\textbf{repo} & The full name of the GitHub repository the commit comes from \\
    \textbf{docfile\_name} & The name of the documentation file. May be useful in the prompt \\
    \rowcolor{gray!30}\textbf{intent} & Small manually gathered intent that describes what we expect from the generated documentation \\
    \textbf{license} & The type of the license in the repository of the commit \\
    \rowcolor{gray!30}\textbf{path\_to\_docfile} & The path to file with documentation in the repository \\
    \textbf{relevant\_code\_files} & List of paths in the repository to the potentially relevant code files \\
    \rowcolor{gray!30}\textbf{relevant\_code\_dir} & Directory with relevant code, field can be empty \\
    \textbf{target\_text} & The text of the target documentation --- ground truth in our task \\
    \rowcolor{gray!30}\textbf{relevant\_code\_context} & Code context joined from relevant code files and directories \\
    \bottomrule
    \end{tabular}
    \end{table}

    \item \textbf{Is there a label or target associated with each instance?}
    \begin{itemize}
        \item The target for each instance is the ground truth documentation text.
    \end{itemize}

    \item \textbf{Is any information missing from individual instances?}
    \begin{itemize}
        \item No.
    \end{itemize}

    \item \textbf{Are relationships between individual instances made explicit?}
    \begin{itemize}
        \item \qXI
    \end{itemize}

    \item \textbf{Are there recommended data splits (\textit{e.g.}, training, development/validation, testing)?}
    \begin{itemize}
        \item \qXII
    \end{itemize}

    \item \textbf{Are there any errors, sources of noise, or redundancies in the dataset?}
    \begin{itemize}
        \item See collection steps in \ref{m2t:data_getting}.
    \end{itemize}

    \item \textbf{Is the dataset self-contained, or does it link to or otherwise rely on external resources?}
    \begin{itemize}
        \item The dataset is self-contained, as it provides the snapshots of all associated repositories.
    \end{itemize}

    \item \textbf{
    Does the dataset contain data that might be considered confidential?
    }
    \begin{itemize}
        \item \qXV
    \end{itemize}

    \item \textbf{Does the dataset contain data that, if viewed directly, might be offensive, insulting, threatening, or might otherwise cause anxiety?}
    \begin{itemize}
        \item \qXVI
    \end{itemize}

    \item \textbf{Does the dataset relate to people?}

    \begin{itemize}
        \item \qXVII
    \end{itemize}

    \item \textbf{Does the dataset identify any subpopulations (\textit{e.g.}, by age, gender)?}
    \begin{itemize}
        \item \qXVIII
    \end{itemize}

    \item \textbf{Is it possible to identify individuals (\textit{i.e.}, one or more natural persons), either directly or indirectly (\textit{i.e.}, in combination with other data) from the dataset?}
    \begin{itemize}
        \item \qXIX
    \end{itemize}

    \item \textbf{Does the dataset contain data that might be considered sensitive in any way?}
    \begin{itemize}
        \item \qXX
    \end{itemize}

    \item \textbf{Any other comments?} 
    \begin{itemize}
        \item No.
    \end{itemize}
\end{enumerate}

\subsection{Collection}

\begin{enumerate}[start=22,label={Q\arabic*}]
    \item \textbf{How was the data associated with each instance acquired?} \label{m2t:data_getting}
    \begin{itemize}
        \item To collect the data, we use the following protocol:

\begin{enumerate}
\item We start with the Python subset of the common corpus of GitHub repositories. For each repository, we extract documentation files --- files with extensions \texttt{.md}, \texttt{.txt}, and \texttt{.rst}, located in the \texttt{docs} directory of the repository.

\item For each documentation file, we extract the associated source code. To do this, we parse the target documentation and extract names of all code files and directories mentioned in it. If a file does not contain any such mentions, we skip it.

\item To further filter the documentation files, we convert documentation into a plain text format by removing specific Markdown syntax (as well as text between Markdown tags like \textit{code}, \textit{autosummary}, etc.). 
We then ensure that each document contains valuable information and has at least 10 lines of text remaining after cleaning. Since the filtering is quite strict, we believe that only important documents remain after this stage.

\item We perform manual review of the datapoints to ensure that the content contains not only information about the code but also summarizes the entire module or project. After manual review, we leave 216 out of 461 files. Most of the files that we reject contain non-informative text that is not related to code. Also, for each documentation file, we manually specify an intent that the model under evaluation can use during generation. 

\end{enumerate}

\item Manual verification is essential, as our experience with data frequently reveals instances where a docfile lacks useful content or does not provide substantial information in the plain text format, without special extensions that enrich documentation.

\end{itemize}

    \item \textbf{What mechanisms or procedures were used to collect the data (\textit{e.g.}, hardware apparatus or sensor, manual human curation, software program, software API)?}
    \begin{itemize}
        \item The data collection step used GitHub API. Then, we performed manual verification and assessment to select and filter data.
    \end{itemize}
    
    \item \textbf{
    If the dataset is a sample from a larger set, what was the sampling strategy?}
    \begin{itemize}
        \item The dataset is sampled from a larger set of repositories by selecting only repositories with Python as the main language and further filtering as described in \ref{m2t:data_getting}.
    \end{itemize}

    \item \textbf{Who was involved in the data collection process (\textit{e.g.}, students, crowdworkers, contractors) and how were they compensated?}
    \begin{itemize}
        \item \qXXV
    \end{itemize}

    \item \textbf{Over what timeframe was the data collected?}
    \begin{itemize}
        \item The construction of this dataset took place between October 2023 and January 2024.
    \end{itemize}

    \item \textbf{Were any ethical review processes conducted?}
    \begin{itemize}
        \item No.
    \end{itemize}

    \item \textbf{Does the dataset relate to people?}
    \begin{itemize}
        \item \qXXVIII
    \end{itemize}

    \item \textbf{Did you collect the data from the individuals in question directly, or obtain it via third parties or other sources (\textit{e.g.}, websites)?}
    \begin{itemize}
        \item \qXXIX
    \end{itemize}

    \item \textbf{Were the individuals in question notified about the data collection?}
    \begin{itemize}
        \item \qXXX
    \end{itemize}

    \item \textbf{Did the individuals in question consent to the collection and use of their data?}
    \begin{itemize}
        \item \qXXXI
    \end{itemize}

    \item \textbf{If consent was obtained, were the consenting individuals provided with a mechanism to revoke their consent in the future or for certain uses?}
    \begin{itemize}
        \item \qXXXII
    \end{itemize}

    \item \textbf{Has an analysis of the potential impact of the dataset and its use on data subjects been conducted?}
    \begin{itemize}
        \item \qXXXIII
    \end{itemize}

    \item \textbf{Any other comments?} 
    \begin{itemize}
        \item No.
    \end{itemize}
\end{enumerate}

\subsection{Preprocessing / Cleaning / Labeling}

\begin{enumerate}[start=35,label={Q\arabic*}]
    \item \textbf{Was any preprocessing/cleaning/labeling of the data done?}
    \begin{itemize}
        \item The data filtering, preprocessing, and labeling steps are described in the data collection procedure in \ref{m2t:data_getting}.

    \end{itemize}

    \item \textbf{Was the “raw” data saved in addition to the preprocessed/cleaned/labeled data?}
    \begin{itemize}
        \item No.
    \end{itemize}

    \item \textbf{Is the software used to preprocess/clean/label the instances available?}
    \begin{itemize}
        \item \qXXXVII
    \end{itemize}

    \item \textbf{Any other comments?} 
    \begin{itemize}
        \item No.
    \end{itemize}
\end{enumerate}

\subsection{Uses}

\begin{enumerate}[start=39,label={Q\arabic*}]

    \item \textbf{Has the dataset been used for any tasks already?}
    \begin{itemize}
        \item We run several LLMs on the collected module summarization dataset with different length of the relevant code context. 
        To assess the quality of the generated documentation, we introduce a new metric called CompScore that uses LLM (Mistral-7B in our case) as an assessor. CompScore feeds the assessor LLM relevant code and two versions of documentation: the ground truth and the model-generated text. The LLM then evaluates which documentation better explains and fits the code. To mitigate variance and potential ordering effects in model responses, we calculate the probability that the generated documentation is superior by averaging the results of two queries:

\[ \text{CompScore} = \frac{P(\text{pred} \mid \text{LLM}(\text{code}, \text{pred}, \text{gold})) + P(\text{pred} \mid \text{LLM}(\text{code}, \text{gold}, \text{pred}))}{2} \]

To count $P(\text{pred} \mid \text{LLM}(\text{code}, \text{pred}, \text{gold}))$, we follow several steps: 

\begin{enumerate}
\item Construct the prompt and feed it into the assessor LLM (see Figure~\ref{figure:p2k_translation}). 

\begin{figure}[h]
\centering
\vspace{0.1cm}
\begin{BVerbatim}
I have 2 different documentations about {intent}. Decide which 
documentation is better: documentation A or documentation B.
My code: [TRIMMED_CODE_CONTEXT]
Documentation A: [PREDICTED_DOC]
Documentation B: [GROUND_TRUTH_DOC]
Better documentation is documentation 
\end{BVerbatim}
\caption{Prompt for the CompScore metric.}
\vspace{0.1cm}
\label{figure:p2k_translation}
\end{figure}

\item Get logits for the next token being ``A'' and ``B'' ($logit_{A}$ and $logit_{B}$) and  convert them into probabilities: 

$$prob_{A}, prob_{B} = \exp{(log\_softmax([logit_{A}, logit_{B}]))}$$

\item $P(\text{pred} \mid \text{LLM}(\text{code}, \text{pred}, \text{gold})) = prob_{A}$ shows the probabilty that the predicted documentation is better than the original from the perspective of the LLM assessor.
\end{enumerate}

\item For our experiments, we use Mistral-7B-Instruct-v0.2 as LLM assessor. We truncate relevant code up to 6,000 tokens in the prompt for metric computation. We evaluate all the models presented in Table \ref{tab:m2t-bench-res} via OpenAI API or TogetherAI API with the same generation parameters. We use zero temperature and predict up to 2,000 new tokens without any penalties to get deterministic results during generation.  
\Cref{tab:m2t-bench-res} shows the results for all the evaluated LLMs with varying length of available relevant code context.

\begin{table}[t]
    \centering
\caption{CompScore metric in the module summarization benchmark for various LLMs.}
\label{tab:m2t-bench-res}
    \begin{tabular}{lcccc} \toprule 
         \multicolumn{1}{c}{\textbf{Model}} & \textbf{128 tokens} & \textbf{512 tokens} & \textbf{1k tokens} & \textbf{2k tokens}
\\ \midrule 
         Mistral-7B-v0.3 & 35.84 & 39.18 & 41.03 & 46.23
\\ 
        Mixtral-8x7B & 34.63 & 38.48 & 39.96 & 40.89
\\ 
        Mixtral-8x22B & 35.33 & 38.48 & 39.49 & 42.24
\\  
        Llama2-7B & 36.33 & 44.21 & 44.13 & 46.19
\\ 
         Llama2-13B & 40.96 & 47.37 & 46.57 & 48.12
\\ 
        Llama2-70B & 39.78 & 45.97 & 46.37 & 48.24
\\ 
        CodeLlama-7B & 33.02 & 36.88 & 36.49 & 38.06
\\ 
        CodeLlama-70B & 38.36 & 38.74 & 39.76 & 37.23
\\ 
        Llama3-8B & 25.37 & 32.14 & 33.84 & 37.35
\\ 
        Llama3-70B & 24.79 & 30.08 & 33.18 & 36.45
\\ 
         Gemma-2B & 16.43 & 21.04 & 21.85 & 25.38 
\\ 
        Gemma-7B & 24.16 & 28.24 & 30.44 & 33.96
\\ 
         GPT-3.5 & 36.83 & 41.59 & 45.59 & 49.48
\\ 
        \textbf{GPT-4} & \textbf{45.62} & \textbf{52.59} & \textbf{56.22} & \textbf{57.33}

\\ \bottomrule
    \end{tabular}

\end{table}

\item We observe that both increasing the context size and the size of the model leads to higher quality. The GPT4 model outperforms the others, achieving a notable CompScore of 57.33. Interestingly, the CodeLlama and Llama3 models show worse performance than the Llama2 model.

    \end{itemize}

    \item \textbf{Is there a repository that links to any or all papers or systems that use the dataset?}
    \begin{itemize}
        \item \qXL
    \end{itemize}

    \item \textbf{What tasks could the dataset be used for?}
    \begin{itemize}
        \item The dataset can be directly employed for the module summarization task. It might be used for other tasks related to the source code changes.
    \end{itemize}

    \item \textbf{Is there anything about the composition of the dataset or the way it was collected and preprocessed/cleaned/labeled that might impact future uses?}
    \begin{itemize}
        \item \qXLII
    \end{itemize}
    
    \item \textbf{Are there tasks for which the dataset should not be used?}
    \begin{itemize}
        \item \qXLIII
    \end{itemize}

    \item \textbf{Any other comments?}
    \begin{itemize}
        \item No.
    \end{itemize}
\end{enumerate}

\subsection{Distribution}

\begin{enumerate}[start=45,label={Q\arabic*}]
    \item \textbf{Will the dataset be distributed to third parties outside of the entity?}
    \begin{itemize}
        \item \qXLV
    \end{itemize}

    \item \textbf{How will the dataset be distributed? Does the dataset have a digital object identifier (DOI)?}
    \begin{itemize}
        \item The dataset is available through DOI at the HuggingFace Hub: {\url{https://doi.org/10.57967/hf/2515}}.
    \end{itemize}

    \item \textbf{When will the dataset be distributed?}
    \begin{itemize}
        \item \qXLVII
    \end{itemize}

    \item \textbf{Will the dataset be distributed under a copyright or other intellectual property (IP) license, and/or under applicable terms of use (ToU)?}
    \begin{itemize}
        \item \qXLVIII
    \end{itemize}

    \item \textbf{Have any third parties imposed IP-based or other restrictions on the data associated with the instances?}
    \begin{itemize}
        \item No.
    \end{itemize}

    \item \textbf{Do any export controls or other regulatory restrictions apply to the dataset or to individual instances?}
    \begin{itemize}
        \item \qL
    \end{itemize}

    \item \textbf{Any other comments?}
    \begin{itemize}
        \item No.
    \end{itemize}
\end{enumerate}

\subsection{Maintenance}
\begin{enumerate}[start=52,label={Q\arabic*}]
    \item \textbf{Who is supporting/hosting/maintaining the dataset?}
    \begin{itemize}
        \item \qLII
    \end{itemize}

    \item \textbf{How can the owner/curator/manager of the dataset be contacted (\textit{e.g.}, email address)?}
    \begin{itemize}
        \item \qLIII
    \end{itemize}

    \item \textbf{Is there an erratum?}
    \begin{itemize}
        \item \qLIV
    \end{itemize}

    \item \textbf{Will the dataset be updated? (\textit{e.g.}, to correct labeling errors, add new instances, delete instances)?}
    \begin{itemize}
        \item \qLV
    \end{itemize}

    \item \textbf{If the dataset relates to people, are there applicable limits on the retention of the data associated with the instances?}
    \begin{itemize}
        \item \qLVI
    \end{itemize}

    \item \textbf{Will older versions of the dataset continue to be supported/hosted/maintained?}
    \begin{itemize}
        \item \qLVII
    \end{itemize}

    \item \textbf{If others want to extend/augment/build on/contribute to the dataset, is there a mechanism for them to do so?}
    \begin{itemize}
        \item \qLVIII 
    \end{itemize}

    \item \textbf{Any other comments?} 
    \begin{itemize}
        \item No.
    \end{itemize}
\end{enumerate}

\end{document}